\documentclass[lettersize,journal]{IEEEtran}
\usepackage{amsmath,amsfonts}
\usepackage{algorithmic}
\usepackage{algorithm}
\usepackage{subfloat}

\usepackage{array}
\usepackage{textcomp}
\usepackage{stfloats}
\usepackage{url}
\usepackage{verbatim}
\usepackage{graphicx}
\usepackage{cite}
\usepackage{amsmath}
\usepackage{amssymb}
\usepackage{bbm}
\usepackage{algorithmic}
\usepackage{wrapfig}
\usepackage{booktabs}

\usepackage{bbm}
\usepackage{bm}
\usepackage{subcaption}
\usepackage{pifont}
\usepackage{multirow}
\usepackage{CJK}
\usepackage{algorithm}
\usepackage{bbding}
\usepackage{indentfirst}
\usepackage{subfloat}

\usepackage{color}
\usepackage{colortbl}
\usepackage{arydshln}
\usepackage{pifont}

\begin{document}

\title{Semantic-Aware Graph Matching Mechanism for Multi-Label Image Recognition}

\author{Yanan Wu, Songhe Feng and Yang Wang
\thanks{Y. Wu, S. Feng are with the Key Laboratory of Big Data \& Artificial Intelligence in Transportation, Ministry of Education, Beijing Jiaotong University, Beijing 100044, China, and also with the School of Computer and Information Technology, Beijing Jiaotong University, Beijing, 100044, China. Y. Wang is with the department of Computer Science, University of Manitoba, Winnipe, MB R3T 2N2, Canada.}
\thanks{Email:\{ynwu0510, shfeng\}@bjtu.edu.cn, ywang@cs.umanitoba.ca}
\thanks{Corresponding author: Songhe Feng}}


\IEEEpubid{\begin{minipage}{\textwidth}\ \\[12pt] \centering
  Copyright © 2023 IEEE. Personal use of this material is permitted. \\
  However, permission to use this material for any other purposes must be obtained from the IEEE by sending an email to pubs-permissions@ieee.org.
\end{minipage}}

\maketitle

\begin{abstract}
Multi-label image recognition aims to predict a set of labels that present in an image. The key to deal with such problem is to mine the associations between image contents and labels, and further obtain the correct assignments between images and their labels. In this paper, we treat each image as a bag of instances, and formulate the task of multi-label image recognition as an instance-label matching selection problem. To model such problem, we propose an innovative Semantic-aware Graph Matching framework for Multi-Label image recognition (ML-SGM), in which Graph Matching mechanism is introduced owing to its good performance of excavating the instance and label relationship. The framework explicitly establishes category correlations and instance-label correspondences by modeling the relation among content-aware (instance) and semantic-aware (label) category representations, to facilitate multi-label image understanding and reduce the dependency of large amounts of training samples for each category. Specifically, we first construct an instance spatial graph and a label semantic graph respectively and then incorporate them into a constructed assignment graph by connecting each instance to all labels. Subsequently, the graph network block is adopted to aggregate and update all nodes and edges state on the assignment graph to form structured representations for each instance and label. Our network finally derives a prediction score for each instance-label correspondence and optimizes such correspondence with a weighted cross-entropy loss. Empirical results conducted on generic multi-label image recognition demonstrate the superiority of our proposed method. Moreover, the proposed method also shows advantages in multi-label recognition with partial labels and multi-label few-shot learning, as well as outperforms current state-of-the-art methods with a clear margin. Our code is available at \textit{https://github.com/yananwu0510/ML-SGM}.
\end{abstract}


\section{Introduction}
Assigning multiple labels to a given image based on its content is known as multi-label image recognition (MLIR). As a fundamental yet essential task in computer vision, it can serve as a prerequisite for many applications, such as weakly supervised localization and segmentation \cite{durand2017wildcat,zhao2019multi}, attribute recognition \cite{wu2020person}, scene understanding \cite{shao2015deeply}, medical diagnosis systems\cite{mahmood2021hybrid},\cite{avola2021multimodal} etc. Compared with single-label image recognition, MLIR is more general and practical since an arbitrary image is likely to contain multiple objects in the physical world \cite{zhou2021multi}. The accurate parsing of multi-label recognition still faces great challenges because of the rich semantic information and complex dependencies of an image and its labels.

\begin{figure}[t]
\centering
\centerline{\includegraphics[scale=0.65]{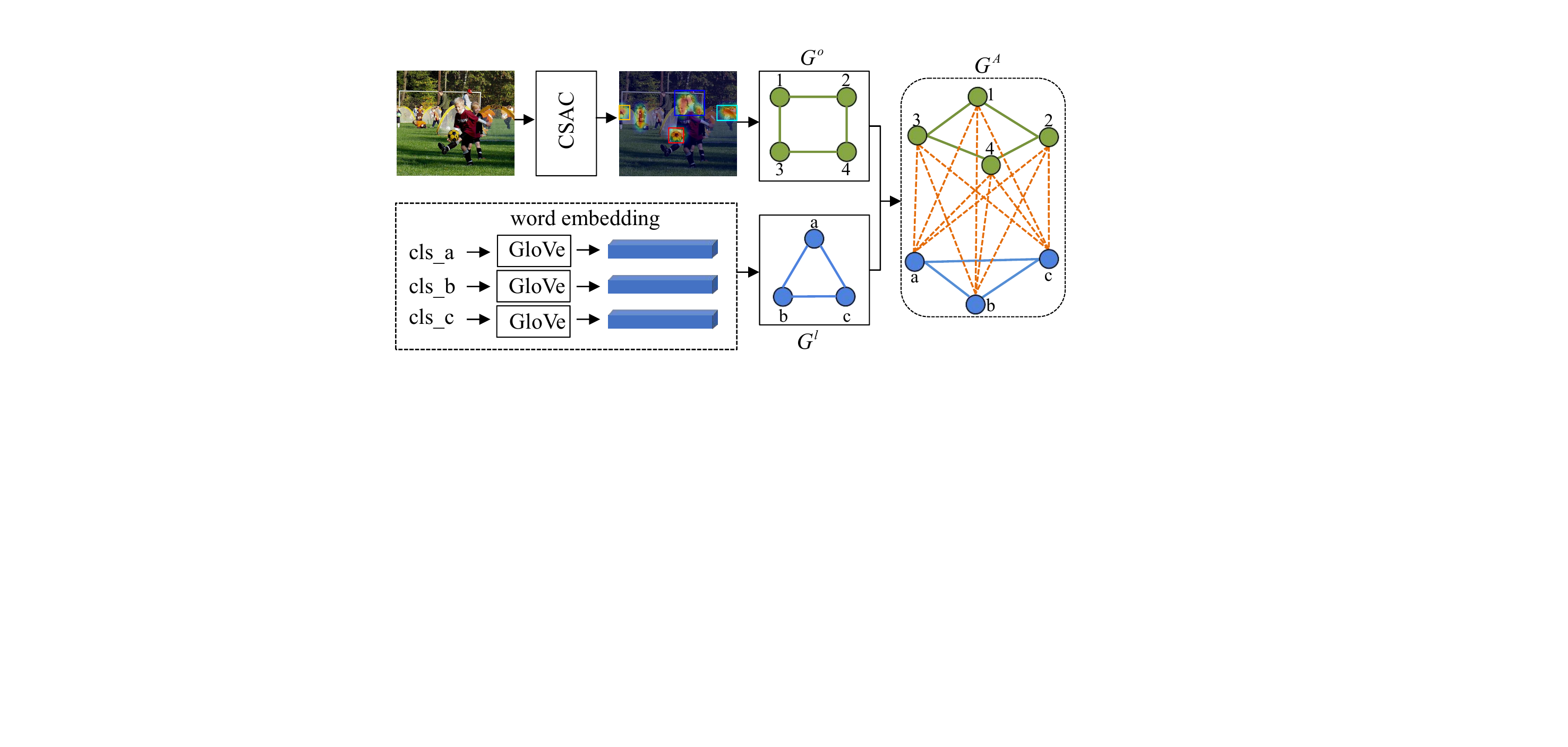}}
\caption{Illustration of the \textbf{AS}signment \textbf{G}raph \textbf{C}onstruction (\textbf{ASGC}). The upper part designs an instance spatial graph $\mathbb{G}^o$, while the lower part builds a label semantic graph $\mathbb{G}^l$. Then each instance is connected to all labels to form the final assignment graph $\mathbb{G}^A$.}
\label{ASGC}
\end{figure}

\IEEEpubidadjcol
The key to accomplish the task of MLIR is how to effectively explore the valuable semantic information from the image context, and further obtain the correct assignments between images and their labels. A simple and straightforward way is to treat each image as a bag of instances/proposals, and cope with each instance in isolation. Thus, the multi-label problem is converted into a set of binary classification problems. However, its performance is essentially limited due to ignoring the complex topology structure among labels. This stimulates research for approaches to capture and mine the label correlations in various ways. For example, Recurrent Neural Networks (RNNs) \cite{wang2016cnn,wang2017multi} and Graph Convolution Network (GCN) \cite{chen2019multi,wang2020multi} are widely used in many MLIR frameworks owing to their competitive performance on explicitly modeling label dependencies. However, most of these methods ignore the associations between semantic labels and image local features, and the spatial contexts of images are not sufficiently exploited. 
Some other works introduce attention mechanisms \cite{chen2020knowledge,you2020cross} to adaptively search semantic-aware instance regions and construct instance-wise correlations based on statistical co-occurrences or semantic embeddings. For example, Chen \textit{et al.} \cite{chen2019learning} propose to incorporate category semantics to better learn instance features and explore their interactions under the guidance of statistical label dependencies. However, this static dependency neglects the characteristics of each image, leading to the hallucination for objects with high-frequent co-occurrences (\textit{e.g.}, hallucinate ``person'' even in a scene containing only ``car''), or negative optimization for objects with less-frequent co-occurrences. Furthermore, due to the lack of fine-grained supervision information, these methods could merely locate instance regions roughly, which neither consider the dynamic interactions among instances nor explicitly describe the instance-label assignment relationship.         

To address the above mentioned issues, in this paper, we fully explore the correspondence (matching) between each instance and label, and reformulate the task of MLIR as an instance-label matching selection problem. Accordingly, we propose an end-to-end deep learning model named Semantic-aware Graph Matching mechanism for Multi-Label image recognition (ML-SGM), which simultaneously incorporates instance spatial relationship, label semantic correlation and the co-occurrence possibility of varying instance-label assignments into a unified framework. Specifically, inspired by Graph Matching (GM) suitable for structured data, we first design a dynamic instance spatial graph by representing each instance feature as the node attributes and the relative location relationship of adjacent instances as the edge attributes. Meanwhile, a label semantic graph is employed to capture the overall semantic correlations, which takes the word embedding of each label as node attributes, and concatenates the attributes of two nodes (labels) associated with the same edge to form the edge attributes. Then each instance is connected to all labels to form the final assignment graph, as shown in Figure \ref{ASGC}, which explicitly models the instance-label matching possibility. Moreover, Graph Network Block (GNB) is introduced to our framework to perform computation on the constructed assignment graph, which forms structured representations for each instance and label by a graph propagation mechanism. Finally, we design a weighted cross-entropy loss to optimize our network output, which indicates the prediction score of each instance-label correspondence.

It is notable that our constructed instance spatial graph can dynamically capture the content-aware category relations for each image. Different from existing methods that extract a corresponding content-aware category representation for each label, our proposed method just generates a small number of image-special instance features, which can reduce the computational complexity of our model. Meanwhile, we model the relationship between instances by their relative positions instead of statistical co-occurrence relationships, which can further enhance the representative and discriminative capabilities of instance features in this adaptive way. In addition, we originally designed our architecture to tackle the MLIR problem, but we would also like to highlight that the proposed ML-SGM is much more general and can be applied for multi-label analysis with limited training samples or annotation information, \textit{e.g.}, multi-label few-shot learning (ML-FSL) \cite{alfassy2019laso,chen2020knowledge} and multi-label recognition with partial labels (MLR-PL) \cite{durand2019learning,huynh2020interactive}. Specifically, our proposed instance spatial graph and label semantic graph can infer the semantic correlation between novel categories and base categories to classify multiple labels using only a few annotated samples effectively. In addition, the proposed instance-label matching module can learn category-special feature similarities and help complement missing labels with high similarities. We conduct extensive experiments on both MLR-PL and ML-FSL tasks and empirical results on various image datasets again demonstrate the advantage of the proposed model. In summary, the main contributions are as follows:       

\begin{itemize}
\item 
To the best of our knowledge, it is the first time to formulate the task of MLIR as a matching selection problem, and accordingly we propose a novel semantic-aware graph matching approach to solve it, where instance spatial relationship, label semantic correlation and co-occurrence possibility of varying instance-label assignments are jointly integrated to guide the correct correspondences between images and their labels.

\item 
Different from previous approaches that build a static statistics-based instance graph, we introduce a dynamic instance spatial graph constructed from content-aware category representations, which can capture the category relations for a specific image in an adaptive way, further enhancing its representative and discriminative ability.

\item We conduct comprehensive experiments on various image recognition tasks to evaluate the effectiveness of the proposed model, including generic MLIR, MLR-PL and ML-FSL tasks. Extensive experimental results have demonstrated that our model exhibits substantial improvements over state-of-the-art methods.    
\end{itemize}

This manuscript is an extension of our IJCAI2021 conference version \cite{wu2021gm}, which differs from \cite{wu2021gm} with the following significant additional contributions. First, instead of using object detection network (Faster-RCNN) \cite{ren2016faster} for instances generation, our proposed ML-SGM directly decomposes the feature map extracted by a CNN backbone into content-aware category representations.
This simplifies the whole framework and saves massive computational costs. Second, we generalize the proposed framework to two more challenging learning tasks, namely \textit{multi-label learning with partial labels} and \textit{multi-label few-shot learning}. We demonstrate the superiority of our approach to these tasks. Finally, we conduct a comprehensive ablation study on several widely used benchmarks and give detailed experimental analysis to demonstrate the effectiveness of each component in our proposed model. 

\begin{table*}[htbp]
\caption{Some important abbreviations and notations.}
\begin{center}
\setlength{\tabcolsep}{4.0mm}{
\begin{tabular}{l|l}
\toprule[1pt]
\textbf{Abbreviation and Notation} &\textbf{Definition} \\\hline\hline
MLIR  &\textbf{M}ulti-\textbf{L}abel \textbf{I}mage \textbf{R}ecognition\\
MLR-PL  &\textbf{M}ulti-\textbf{L}abel \textbf{R}ecognition with \textbf{P}artial \textbf{L}abels\\
ML-FSL &\textbf{M}ulti-\textbf{L}abel \textbf{F}ew-\textbf{S}hot \textbf{L}earning \\
ML-SGM &\textbf{S}emantic-aware \textbf{G}raph \textbf{M}atching framework for \textbf{M}ulti-\textbf{L}abel image recognition \\
ASGC &  \textbf{AS}signment \textbf{G}raph \textbf{C}onstruction module\\
CSAC & \textbf{C}ontent-aware \textbf{S}emantic \textbf{AC}tivation module\\
ILMS &\textbf{I}nstance-\textbf{L}abel \textbf{M}atching \textbf{S}election module\\ \hline
$\mathcal{D}=\{(\textbf{X}_i,\textbf{y}_i)\}_{i=1}^N$ & multi-label image dataset \\
${\textbf{X}_i}=[\textbf{x}_i^0,\textbf{x}_i^1,...,\textbf{x}_i^M]^T\in{\mathbb{R}^{(M+1)\times D}}$ &$i$-th image that consists of $M+1$ instances \\
$\textbf{y}_i=[y_i^1,y_i^2,...,y_i^C]^T$ & ground-truth label vector of $\textbf{X}_i$ \\
$\textbf{x}_i^0$; $\textbf{x}_i^1 \sim \textbf{x}_i^M$ & whole image feature; semantic-aware instance features\\
$\textbf{x}_i^j \rightarrow \textbf{x}^j$ &simplify  $\textbf{x}_i^j$ by $\textbf{x}^j$ to represent a specific instance feature for the convenience of description \\
$\mathbb{G}^o$; $\mathbb{G}^l$; $\mathbb{G}^A$ &instance
spatial graph; label semantic graph; instance-label assignment graph\\
$\varphi_{enc}^{v}$; $\varphi_{enc}^{e}$; $\varphi_{dec}^{e}$ &encoder of node attributes; encoder of edge attributes; decoder of edge attributes\\
$\mathbb{N}^{o}$; $\mathbb{N}^{l}$ &the sets of instance nodes and label nodes adjacent with current node \\
\bottomrule[1pt]
\end{tabular}}
\label{notations}
\end{center}
\vspace{-3mm}
\end{table*}

\section{Related Works }
\subsection{Multi-Label Image Recognition}
The task of multi-label image recognition (MLIR) has attracted increasing interest recently. A straightforward way to address this problem is to train independent binary classifiers for each label. However, such method does not consider the relationships among labels, and the number of predicted labels will grow exponentially as the number of categories increases. To overcome the challenge of such an enormous output space, some works convert the multi-label problem into a set of multi-class problems over region proposals. Wei \textit{et al.} \cite{wei2015hcp} extracted an arbitrary number of object proposals, then aggregated the label confidences of these proposals with max-pooling to obtain the final multi-label predictions. Yang \textit{et al.} \cite{yang2016exploit} treated each image as a bag of instances, and solved the MLIR task in a multi-instance learning manner. However, the above methods ignore the label correlations in multi-label images when converting MLIR to the multi-class task. 

Recently, researchers have focused on exploiting the label correlation to facilitate the learning process \cite{markatopoulou2018implicit}, \cite{wang2021semantic}, \cite{yan2018participation}. Gong \textit{et al.} \cite{gong2013deep} leveraged a ranking-based learning strategy to train deep convolutional neural networks for MLIR and found that the weighted approximated-ranking loss can implicitly model label correlation and achieve state-of-the-art performance. Wang \textit{et al.} \cite{wang2016cnn} combined recurrent neural networks with CNNs to capture the correlations between labels and predicted labels in a predefined order.
Besides, some studies \cite{zang2022multi}, \cite{yan2020higcin} utilized graph structure to model the label correlation. Chen \textit{et al.} \cite{chen2019multi} used GCN to map a group of label semantic embeddings into inter-dependent classifiers.     
Wang \textit{et al.} \cite{wang2020multi} proposed to model label correlation by superimposing label graph built from statistical co-occurrence information into the graph constructed from knowledge priors of labels. However, none of the aforementioned methods consider the associations between semantic labels and image local features, and the spatial contexts of images have also not been sufficiently exploited.

To solve the above issues, recent progress on MLIR attempts to model label correlation with region-based multi-label approaches. Zhu \textit{et al.} \cite{zhu2017learning} designed a spatial regularization network to explore both spatial and semantic relationships among multiple labels according to weighted attention maps.
Chen \textit{et al.} \cite{chen2019learning} incorporated category semantics to better learn instance features and explored their interactions under the guidance of statistical label dependencies. 
Ye \textit{et al.} \cite{ye2020attention} proposed an Attention-Driven Dynamic Graph Convolutional Network, which consists of a static graph (label statistics) for capturing the global coarse category dependencies and a dynamic graph for exploiting content-dependent category relations, respectively. Our work is largely inspired by these region and GCN-based multi-label methods. However, instead of extracting a corresponding content-aware representation across all categories for each specific-image, our ML-SGM just generates a small number of image-special instance features, which can reduce the computational complexity. And we model the relations between instances by their relative positions instead of statistical label co-occurrence \cite{chen2020knowledge} or instance feature concatenation \cite{ye2020attention}, which can further enhance the representative and discriminative of categories representation.    

\subsection{Multi-Label Recognition with Partial Labels}
Multi-label tasks often involve incomplete training data, hence several methods \cite{huynh2020interactive} have been proposed to solve the problem of multi-label learning with partial labels, \textit{i.e.}, merely some labels are known while other labels are missing (also called unknown labels). In general, existing works to deal with this issue are roughly grouped into four categories. (1) Missing labels are directly treated as negative labels \cite{sun2017revisiting,mahajan2018exploring}. Common to these methods is that the label bias is brought into the objective function. As a result, their performance suffers from a severe drop when massive ground-truth positive labels are wrongly annotated as negative. (2) Missing labels are filled via a matrix completion scheme \cite{wu2018multi}. These methods utilize instance-level similarity and label co-occurrence with low-rank regularization on the label matrix to complete missing labels.
Wu \textit{et al.} \cite{wu2015ml} exploited a mixed graph to encode a network of label dependencies to transfer the given labels knowledge into missing labels. (3) Missing labels are treated as latent variables in probabilistic models and updated in an active learning manner. Kapoor \textit{et al.} \cite{kapoor2012multilabel} proposed to project the label vector to a lower dimensional space and predict missing labels by posterior inference using Bayesian networks.  
(4) Unlike most of the previous methods that treat missing labels as negative labels, some other works formulate it as the third label state to explicitly distinguish label information in partially labeled data. For example, Wu \textit{et al.} \cite{wu2014multi} defined three label states, including positive labels $+1$, negative labels $-1$ and missing labels $0$, to recover the full label assignment for each sample by enforcing the consistency with available label assignments and the smoothness of label assignments.            

Recently, some CNN-based deep methods have been proposed to tackle the MLIR with missing labels problem, which are scalable and end-to-end learnable. Durand \textit{et al.} \cite{durand2019learning} designed a normalized binary cross-entropy (BCE) loss to exploit label proportion information and use it to train the model with partial labels. 
Huynh \textit{et al.} \cite{huynh2020interactive} introduced statistical label co-occurrence and image-level feature similarity to regularize training networks to avoid overfitting to partial labels. Chen \textit{et al.} \cite{chen2022structured} explored semantic correlations to transfer knowledge of known labels to generate pseudo labels for missing labels and use both known and generated labels for model training. In this paper, we further develop our framework that allows efficient end-to-end training with partial labels.

\subsection{Few-Shot Learning}
Few-shot learning aims to recognize new objects from extremely limited training samples, which learns from the base dataset with many labeled training samples and can be generalized
to novel dataset with few labeled samples \cite{wertheimer2021few}, \cite{tang2022learning}. 
The key to solve this problem is to transfer the knowledge of the base categories to the novel categories, further achieving good generalization \cite{yazdanpanah2022revisiting,shao2021mhfc,xu2022gct}. 
In the standard formulation of few-shot learning, each few-shot task has a \textit{support set} and a \textit{query set}. Let $\mathcal{S}$ denote the support set, which contains $N$ different image classes and $K$ labeled samples per class (called $N$-way $K$-shot setting). Given the query set $\mathcal{Q}$, few-shot learning aims to classify each unlabeled sample in $\mathcal{Q}$ according to the set $\mathcal{S}$.
In summary, existing few-shot methods can be grouped into the following aspects.

Metric learning-based methods put emphasis on finding better distance metrics to determine the correct categories for the query samples \cite{jiang2020multi}. MatchingNet \cite{vinyals2016matching} learned a sample-wise metric that maps a small labeled support set and an unlabelled sample to its label, generating the labels of the query set in terms of distances between samples. ProtoNet \cite{snell2017prototypical} extended the samples-wise to class-wise metric, which adopts the mean of the embedded sample features in novel categories as a class prototype and then recognizes test samples by nearest-neighbor classifiers. 

\begin{figure*}[!htbp]
\centering  
\centerline{\includegraphics[scale=0.65]{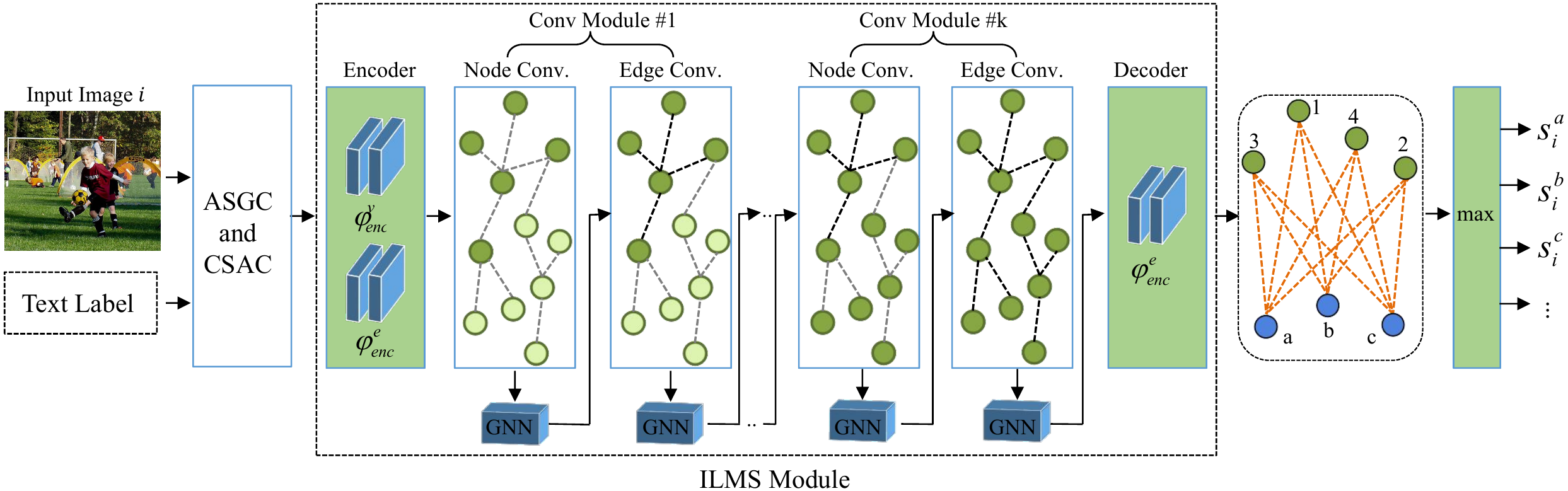}}
\caption{Illustration of our proposed deep learning framework for MLIR task. Overall our model consists of three major components: the \textbf{C}ontent-aware \textbf{S}emantic \textbf{AC}tivation (\textbf{CSAC}), the \textbf{AS}signment \textbf{G}raph \textbf{C}onstruction (\textbf{ASGC}) and the \textbf{I}nstance-\textbf{L}abel \textbf{M}atching \textbf{S}election (\textbf{ILMS}). The CSAC module is contained in the ASGC module, and the details of ASGC and CSAC are shown in Figure \ref{ASGC} and \ref{CSAC}, respectively. The ILMS module consists of Encoder, Graph Convolution and Decoder. The encoder and decoder are designed as MLPs, where encoder transforms all node attributes and edge attributes into latent space and decoder derives instance-label matching scores from the updated graph state. Moreover, the graph convolution utilizes Graph Network Block (GNB) to perform nodes and edges attributes aggregating and updating.}
\label{overview}
\end{figure*}

Meta-learning-based methods follow the key idea of learning to learn that optimizes the models for the capacity of rapidly adapting to new categories \cite{shao2021mdfm}, \cite{tang2020blockmix}. Early works focused
on meta-learning an optimizer including the LSTM-based meta-learner \cite{ravi2016optimization} and the external-memory assisted weight updating \cite{munkhdalai2017meta}. Later, researches concentrated on meta-learning a model initialization \cite{rajeswaran2019meta} that is suitable for fast adapting the model to a new class by fine-tuning on a few support samples with few iterations. But such methods are faced with a difficult bi-level optimization problem owing to the inter-dependency of the parameters updated in the inner and outer loops for each episode. To solve this issue, Lee \textit{et al.} \cite{lee2019meta} used implicit differentiation for quadratic programs in a final SVM layer and Sun \textit{et al.} \cite{sun2019meta} proposed to learn task-relevant scaling and shifting functions to dynamically adjust the CNN weights.

Data augmentation-based methods aim to synthesize new samples from the few training samples to augment the generalization of the model. In addition to simple data augmentation tricks such as horizontal flipping, scaling and positional shifts, more advanced approaches are proposed to learn compositional representations for novel categories recognition. Schwartz \textit{et al.} \cite{schwartz2018delta} proposed a delta-encoder to extract transferable intra-class deformations and synthesize samples for novel categories. Alfassy \textit{et al.} \cite{alfassy2019laso} focused on a new and challenging problem - multi-label zero-shot learning, and proposed a label-set operation network LaSO to synthesize samples with multiple labels. In this paper, we generalize our model to this task and demonstrate its superior ability to learn robust category features and classifiers from limited samples.

\section{The Proposed Method} 
Given a multi-label image dataset $\mathcal{D}=\{(\textbf{X}_i,\textbf{y}_i)\}_{i=1}^N$, we denote ${\textbf{X}_i}=[\textbf{x}_i^0,\textbf{x}_i^1,...,\textbf{x}_i^M]^T\in{\mathbb{R}^{(M+1)\times D}}$ as the $i$-th image that consists of $M+1$ instances, $\textbf{y}_i=[y_i^1,y_i^2,...,y_i^C]^T$ as the ground-truth label vector for $\textbf{X}_i$, where each instance $\textbf{x}_i^j (j=0,1,...,M)$ is a $D$-dimensional feature vector (here $\textbf{x}_i^0$ denotes the whole image and $\textbf{x}_i^1 \sim \textbf{x}_i^M$ denote the semantic-aware instances). $C$ is the number of all possible labels in the dataset. ${y_i^c}=1$ indicates that image $\textbf{X}_i$ is annotated with label $c(c=1,2,...,C)$, and ${y_i^c}=0$ otherwise. The goal of ML-SGM is to learn a multi-label classification model from the instance-level feature together with the image-level ground-truth label, and further assign the predictive labels for a test image. For convenience, we list some crucial abbreviations and notations in Table \ref{notations}. 

\subsection{Overview}
The overview architecture of the proposed ML-SGM is illustrated in Figure \ref{overview}, which consists of three components: the Content-aware Semantic ACtivation (CSAC), the ASsignment Graph Construction (ASGC) and the Instance-Label Matching Selection (ILMS). The CSAC module dynamically generates a small number of instance regions while keeping their diversity as high as possible. The ASGC module takes instance spatial graph and label semantic graph as input, and considers each instance-label connection as a candidate matching edge. The ILMS module is another core component of our learning framework, which introduces Graph Network Block (GNB) to convolve the information of neighborhoods of each instance and label to form a structured representation through several convolution operators. Finally, our model derives a prediction score for each instance-label correspondence and optimizes such correspondence with a weighted cross-entropy loss.   

\begin{figure*}[!htbp]
\centering  
\centerline{\includegraphics[scale=0.85]{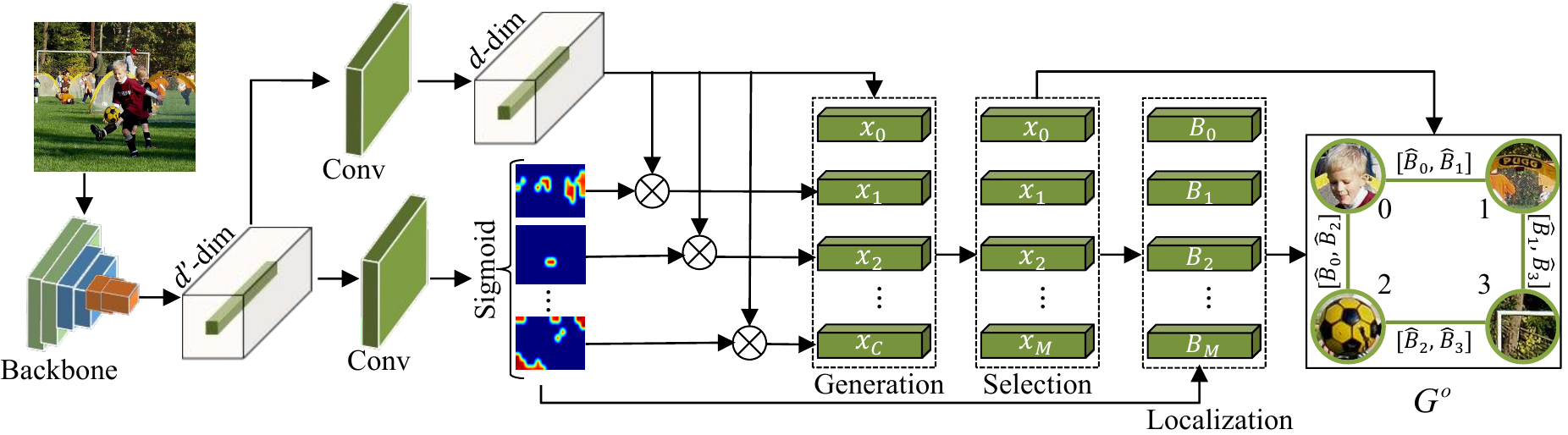}}
\caption{Illustration of the Content-aware Semantic ACtivation (CSAC). Different from the semantic attention module in ADD-GCN \cite{ye2020attention} that extracts a corresponding content-aware representation across all categories for each image, our CSAC generates a small number of instance regions while keeping their diversity and computational efficiency. We model the relations between instances by their relative positions instead of the concatenation of instance features. This can further enhance the representative and discriminative capabilities of category representations.}
\label{CSAC}
\vspace{-3mm}
\end{figure*}

\subsection{The Content-aware Semantic Activation}
In MLIR task, visual concepts are highly related with the local regions of an image \cite{huynh2020shared}. In order to better mine the information of local regions and further explore their interplay, the instance regions should be generated in an efficient manner. The desirable generation module and instance regions need to satisfy the following basic principles, including \textit{high diversity}, \textit{small number of instance regions} and \textit{high computational efficiency}. Following these principles, we introduce the Content-aware Semantic ACtivation module (CSAC) to dynamically obtain image-specific instance representations (Figure \ref{CSAC}), which includes three steps: class-activation generation, class selection and instance localization.  

\paragraph{\textbf{Class-Activation Generation}} In our model, we apply class activation
mapping (CAM) \cite{zhou2016learning} to generate instance regions. Specifically, we feed an image to a pre-trained backbone network \cite{he2016deep} to extract its feature maps $\textbf{F} \in \mathbb{R}^{D\times H\times W}$ from the last convolutional layer, where $D$, $H$, $W$ are the number of channels, height, and width of the feature maps, respectively. Then a new convolution layer as the classifier is performed on the feature maps $\textbf{F}$ to generate the content-aware semantic activation map $\textbf{M}\in \mathbb{R}^{C\times W \times H}$, and regularize it via a \textit{sigmoid} function. It is worth noting that these operations are different from CAM, but have better performance in our experiments.    
Next, the semantic activation map $\textbf{M}$ is used to convert the feature map $\textbf{F}$ into a set of instance representations\footnote{For the convenience of description, in the subsequent content, we simplify  $\textbf{x}_i^j$ by $\textbf{x}^j$ to represent a specific instance feature.}, \textit{i.e.}, $\overline{\textbf{X}}=\{\textbf{x}^1,\textbf{x}^2,...,\textbf{x}^C\}$. Each instance representation $\textbf{x}^c$  can selectively aggregate features related to its specific category $c$ as follows, 
\begin{equation}
\textbf{x}^c=\sum_{x=1}^{H}\sum_{y=1}^{W}\textbf{M}_{c}(x,y)\textbf{F}(x,y).
\end{equation}

The discriminative instance regions of a specific $\textbf{x}^c$ are significantly different among all possible instance representations $\{\textbf{x}^c\}_{c=1}^C$. If we use $\{\textbf{x}^c\}_{c=1}^C$ to localize the potential instance regions, it is easy to increase the diversity of different instance regions and cover all single objects of the given multi-label image. Furthermore, in order to make instance representation be more tailored for the multi-label data, we fine-tune the representation on our target dataset with the weighted cross-entropy loss. This can help to overcome the localization problem of previous unseen categories as the model has already learned semantic-aware features from a wide range of categories. 

\paragraph{\textbf{Class selection}} The number of the above-generated instance representations is equal to that of all categories associated with the corresponding dataset. For example, there are 20 and 80 categories on PASCAL VOC and MS-COCO datasets, respectively. If we employ all instances, it would cause two issues. (i) The number of generated instances is too large to be computationally efficient. (ii) The majority of instances will be redundant or meaningless since an image is usually composed of several instances. Moreover, it is a reasonable
assumption that high category confidence means that
the corresponding instance is present in the image with a high
probability. Therefore, in our model, we select instances of which the prediction scores are greater than the threshold $\gamma$ as the image visual features $\overline{\textbf{X}}'=[\textbf{x}^1,\textbf{x}^2,...,\textbf{x}^M]^T\in \mathbb{R}^{M\times D}$ to learn image-specific category correlation. Here, the prediction scores are generated by performing the global spatial pooling on the semantic activation map. $M$ is the number of selected instances and the choice of the threshold $\gamma$ is analyzed in Section 4.7. In addition to the instances selected based on the above criterion, we also extract the global features of the whole image as the representation $\textbf{x}^0$ of a new instance, which serves as the complementary information to handle potential missing target instances or abstract concepts, \textit{e.g.}, ``travel'' or ``singing'', etc. The representations of these instances eventually form the image visual features ${\textbf{X}}=[\textbf{x}^0,\textbf{x}^1,...,\textbf{x}^M]^T\in{\mathbb{R}^{(M+1)\times D}}$.    

\begin{figure}[t]
\centering
\centerline{\includegraphics[scale=0.40]{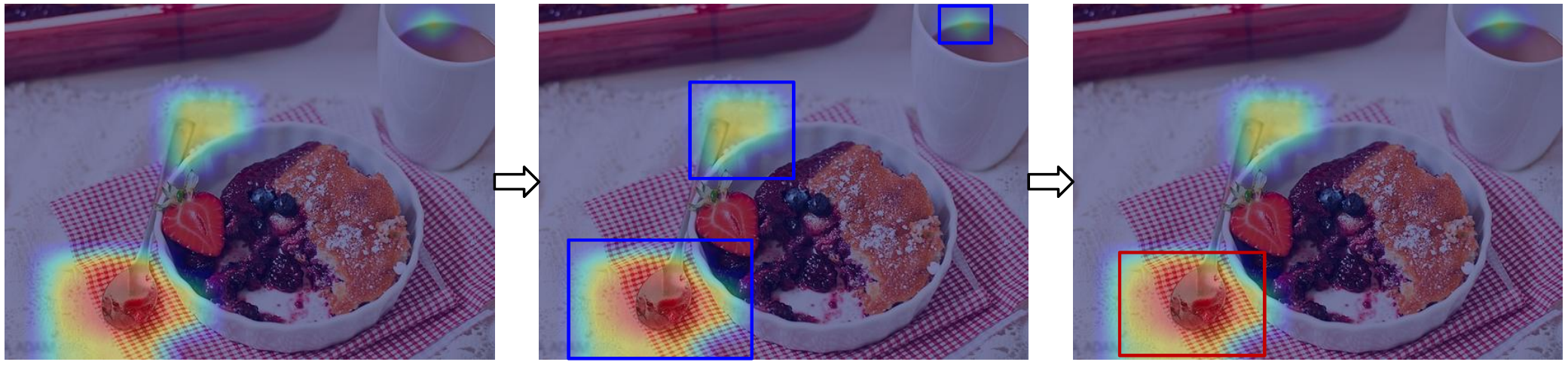}}
\caption{Semantic activation map and instance region localization of the \textit{spoon} category. The \textcolor{blue}{blue} boxes indicate instance region localization before employing the thresholding method. The \textcolor{red}{red} box indicates instance region localization after employing the threshold method.}
\label{instance_loc}
\vspace{-3mm}
\end{figure}

\paragraph{\textbf{Instance localization}}
The value of produced $\textbf{M}_{c}(x,y)$ indicates the importance of the activation at spatial location $(x, y)$ leading to classify an image to category $c$. In this work, our objective is to recognize multi-class objects in a given image, and only one discriminative region needs to be selected for each category. Therefore,  some constraints have to be added such that a unique and important interval among multiple feasible intervals can be chosen.
In order to efficiently localize such instance regions, we exploit a simple thresholding method to segment the heatmap. Specifically, we first segment the regions whose values are above 20\% of the max intensity of the semantic activation map. It results in connected segments of pixels and then we draw the bounding box $\textbf{B}(x,y,w,h)$ around the single largest segment.  As shown in Figure \ref{instance_loc}, if the thresholding method is applied, the visual features of the \textit{spoon} category are represented as the instance regions in the red box, otherwise are represented as all the instance regions in the blue boxes. In addition, such a thresholding method has been widely used in weakly-supervised localization tasks \cite{selvaraju2017grad,zhou2016learning,jiang2021layercam} to generate object bounding boxes from class activation maps. Here 20\% is a hyperparameter that is empirically set.

\subsection{The Assignment Graph Construction}
As depicted in Figure \ref{ASGC}, we first construct an instance spatial graph to explore the relationship of the spatially adjacent instances. Specifically, we feed an image to the CSAC module to generate a set of semantic-aware instances, where each instance contains a bounding box $\textbf{B}(x,y,w,h)$. Within the bounding box, $(x,y)$ denotes the location ($x$-axis, $y$-axis coordinates) and $(w,h)$ denotes the size (width and height). Each of these instances is taken as a node in the instance spatial graph $\mathbb{G}^o$ in which the edges between each pair of nodes are produced through $k$-Nearest Neighbor criteria. Note that our $\mathbb{G}^o$ is a directed graph, where the attributes of nodes are denoted by the visual feature of the corresponding instances, and the attributes of edges are represented by the concatenations of bounding box coordinates of its source instance and target instance
\begin{equation}
\textbf{v}_i^{o}=\textbf{x}^i,
\hspace{1em}
\textbf{e}_{ij}^{o}=[\hat{\textbf{B}}_i,\hat{\textbf{B}}_j],
\end{equation}
where $\textbf{x}^i$ and $\hat{\textbf{B}}_{i}(x_i,y_i,w_i,h_i)$ denote the features and location coordinates of $i$-th instance respectively and $[\cdot,\cdot]$ the concatenations of its input. It is worth mentioning that the concatenation operation has been widely used to fuse the features in many learning paradigms.

Similar to \cite{chen2019multi,you2020cross}, we also construct a label semantic graph $\mathbb{G}^l$ to capture the topological structure in the label space, where each node of the graph is represented as word embedding of the label. However, different from the above methods, our $\mathbb{G}^l$ does not require the pre-defined label co-occurrence statistics. Instead, we form the initial edge attributes of $\mathbb{G}^l$ by combining the word embeddings of the connected two label nodes.     
\begin{equation}
\textbf{v}_i^{l}=\textbf{w}_i,
\hspace{1em}
\textbf{e}_{ij}^{l}=[\textbf{w}_i,\textbf{w}_j],
\end{equation}
where $\textbf{w}_i$ denotes the word embedding of $i$-th label.

To explicitly establish the instance-label matching relationship, we connect each instance in $\mathbb{G}^o$ to all labels in $\mathbb{G}^l$ to form the initial instance-label assignment graph $\mathbb{G}^A$. In $\mathbb{G}^A$, the attributes of the matching edges that connect instances and labels are represented as
\begin{equation}
\textbf{e}_{ij}^{m}=[\textbf{v}_i^{o},\textbf{v}_j^{l}].
\end{equation}
where $\textbf{v}_i^{o}$ is the $i$-th instance in the visual/object space, $\textbf{v}_j^{l}$ denotes the $j$-th label in the label space, and $\textbf{e}_{ij}^{m}$ indicates the attribute of the matching edge between the $i$-th instance and the $j$-th label.

In this way, we successfully convert the problem of building the complex correspondence between an image and its labels to the issue of selecting reliable instance-label matching edges from a constructed assignment graph. Accordingly, the goal of the MLIR problem is transformed into how to solve the matching selection problem and obtain the optimal instance-label assignment.

\subsection{Modeling Instance-Label Correspondence}
As illustrated in Figure \ref{overview}, our matching selection module is designed on the top of Graph Network Block (GNB) presented in \cite{wang2020learning}, which defines a class of functions for relation reasoning over graph-structured representations. It is extended in our model to fit the instance-label matching selection problem by eliminating redundant components and redefining certain functions. Specifically, the matching selection module consists of three main components: Encoder, Graph Convolution and Decoder.  

\paragraph{\textbf{Encoder}} The encoder takes the constructed assignment graph $\mathbb{G}^A$ as input, and transforms its attributes into a latent feature space by two parametric update functions $\varphi_{enc}^{v}$ and $\varphi_{enc}^{e}$. In our framework, $\varphi_{enc}^{v}$ and $\varphi_{enc}^{e}$ are designed as multi-layer perceptrons (MLPs), each of which takes respectively a node attributes vector and an edge attributes vector as input and transforms them into latent spaces. Note that $\varphi_{enc}^{v}$ is mapped across all nodes (instances and label) to compute per-node updates, and $\varphi_{enc}^{e}$ is mapped across all edges (relationships) to compute per-edge updates.

Formally speaking, we denote $\varphi_{enc}^{v}(\mathcal{V}^{A})$ and $\varphi_{enc}^{e}(\mathcal{E}^{A})$ as the updated node attributes and edge attributes by applying $\varphi_{enc}^{v}$ and $\varphi_{enc}^{e}$ to each node and each edge respectively. Then the encoder module can be briefly described as 
\begin{equation}
\mathbb{G}^A \leftarrow encode(\mathbb{G}^A)=(\mathbb{V}^A,\mathbb{E}^A,\varphi_{enc}^{v}(\mathcal{V}^{A}),\varphi_{enc}^{e}(\mathcal{E}^{A})).
\end{equation}

The updated graph $\mathbb{G}^A$ is then passed to the subsequent convolution modules as input.

\paragraph{\textbf{Graph Convolution Module}} This module consists of a node convolution layer and an edge convolution layer. The node convolution layer collects the attributes of all the nodes and edges adjacent to each node to compute per-node updates. It is followed by the edge convolution layer that assembles the attributes of the two nodes associated with each edge to generate new attributes of this edge. 

Specifically, for the $i$-th node in $\mathbb{G}^o$, the aggregation function gathers the information from its adjacent nodes and associated edges, and the update function outputs the updated attributes according to the gathered information. Given that for each instance node $\textbf{v}_i^{o}$, other instance nodes and label nodes are connected with instance edges and matching edges respectively, we design two types of aggregation functions as 
\begin{equation}
\hat{\textbf{v}}_i^{o}=\frac{1}{\|\mathbb{N}^{o}\|}\hat{\rho}_{n}^{o}([\textbf{e}_{ij}^{o},\textbf{v}_j^{o}]),
\hspace{1em}
\tilde{\textbf{v}}_i^{o}=\frac{1}{\|\mathbb{N}^{l}\|}\tilde{\rho}_{n}^{o}([\textbf{e}_{ij}^{c},\textbf{v}_j^{l}]),
\end{equation}
where $\mathbb{N}^{o}$ and $\mathbb{N}^{l}$ are two sets of instance nodes and label nodes adjacent with $\textbf{v}_i^{o}$. $\hat{\rho}_{n}^{o}$ and $\tilde{\rho}_{n}^{o}$ gather the information from object nodes and label nodes respectively.
Then an update function is employed to update the attributes for $\textbf{v}_i^{o}$
\begin{equation}
\textbf{v}_i^{o}\leftarrow \phi^{o}_{n}([\textbf{v}_i^{o},\hat{\textbf{v}}_i^{o},\tilde{\textbf{v}}_i^{o}]),
\end{equation}
where $\phi^{o}_{n}$ takes the concatenation of the current attributes of $\textbf{v}_i^{o}$ and gathered information $\hat{\textbf{v}}_i^{o}$ and $\tilde{\textbf{v}}_i^{o}$, and outputs the updated attributes for $\textbf{v}_i^{o}$.

Similar to instance nodes, the label nodes are connected with two types of nodes and associated with two types of edges. Therefore, we also design two aggregation functions and an update function for label node $\textbf{v}_i^{l}$. Specifically, the aggregation functions are formulated as
\begin{equation}
\hat{\textbf{v}}_i^{l}=\frac{1}{\|\mathbb{N}^{l}\|}\hat{\rho}_{n}^{l}([\textbf{e}_{ij}^{l},\textbf{v}_j^{l}]),
\hspace{1em}
\tilde{\textbf{v}}_i^{l}=\frac{1}{\|\mathbb{N}^{o}\|}\tilde{\rho}_{n}^{o}([\textbf{e}_{ij}^{c},\textbf{v}_j^{o}]),
\end{equation}
and the update function is represented as
\begin{equation}
\textbf{v}_i^{l}\leftarrow \phi^{l}_{n}([\textbf{v}_i^{l},\hat{\textbf{v}}_i^{l},\tilde{\textbf{v}}_i^{l}]).
\end{equation}

For an instance edge $\textbf{e}_{ij}^{o}$, both of its source node and receive node are in $\mathbb{G}^o$. Therefore, we design an aggregation function and the update function as follows
\begin{equation}
\hat{\textbf{e}}_{ij}^{o}=\rho_{e}^{o}([\textbf{v}_i^{o},\textbf{v}_j^{o}]),
\hspace{1em}
{\textbf{e}}_{ij}^{o}\leftarrow \phi^{o}_{e}([\textbf{e}_{ij}^{o},\hat{\textbf{e}}_{ij}^{o}]),
\end{equation}
where $\rho_{e}^{o}$ aggregates the information from source instance node $\textbf{v}_i^{o}$ and receive node $\textbf{v}_j^{o}$, $\phi^{o}_{e}$ updates the attributes of ${\textbf{e}}_{ij}^{o}$ according to the gathered information.

Similar to the instance edge convolution operator, the label edge convolution layer consists of an aggregation function and an update function, which are designed as
\begin{equation}
\hat{\textbf{e}}_{ij}^{l}=\rho_{e}^{l}([\textbf{v}_i^{l},\textbf{v}_j^{l}]),
\hspace{1em}
\textbf{e}_{ij}^{l}\leftarrow \phi^{l}_{e}([\textbf{e}_{ij}^{l},\hat{\textbf{e}}_{ij}^{l}]).
\end{equation}

Different from instance edges and label edges, the matching edges connect instance nodes and label nodes. Therefore, the aggregation function gathers information from different types of nodes
\begin{equation}
\hat{\textbf{e}}_{ij}^{m}=\rho_{e}^{m}([\textbf{v}_i^{o},\textbf{v}_j^{l}]),
\end{equation}
and the update function also takes the combination of aggregated features and its current features as input and outputs the updated attributes
\begin{equation}
\textbf{e}_{ij}^{m}\leftarrow \phi^{m}_{e}([\textbf{e}_{ij}^{m},\hat{\textbf{e}}_{ij}^{m}]).
\end{equation}

All the above aggregation functions and update functions are designed as MLPs, but their structure and parameters are different from each other. 

\paragraph{\textbf{Decoder}} The decoder module reads out the final output from the updated graph state. Since only the attributes of instance-label matching edges are required for final evaluation, the decoder module contains only one update function $\varphi_{dec}^{e}$ that transforms the edges attributes into the desired space
\begin{equation}
S=decode(\mathbb{G}^A)=\varphi_{dec}^{e}(\mathcal{E}^{A}),
\end{equation}
where $S\in {[0,1]}^{(M+1)\times C}$ denotes the prediction score that each instance is matched with the corresponding label. Similarly, $\varphi_{enc}^{e}$ is parameterized by an MLP. 

\subsection{Optimizing Multi-Label Prediction}
In order to interpret each ground truth label of the input image, there should be at least one instance that best matches it. With consideration of the existence of noisy instances, a cross-instance max-pooling is carried out to fuse the output of our framework into an integrative prediction. Suppose $\textbf{s}_{j}(j=0,1,...,M)$ is the prediction score vector of the $j$-th instance from the decoder and $s_j^c(c=1,2,...,C)$ is the $c$-th category matching score of $s_j$. The cross-instance max-pooling can be formulated as 
\begin{equation}
p_i^c=max(s_0^c,s_1^c,...,s_M^c),
\end{equation}
where $p_i^c$ can be considered as the prediction score for the $c$-th category of the given image $i$. 

\paragraph{\textbf{Multi-Label Image Recognition}} For generic MLIR task, the training process of our network is guided by a weighted cross-entropy loss with the ground-truth labels $\textbf{y}_i$ as supervision,
\begin{gather}
 \mathcal{L}=\sum_{i=1}^{N}\sum_{c=1}^{C}w^c[y_i^c\log(p_i^c)+(1-y_i^c)\log(1-p_i^c))]\nonumber\\
 w^c=y_i^c\cdot e^{\beta(1-r^c)}+(1-y_i^c)\cdot e^{\beta r^c}, 
 \label{loss}
 \end{gather}
where $w^c$ is used to alleviate the class imbalance, $\beta$ is a hyper-parameter and $r^c$ is the ratio of label $c$ in the training set. 

\paragraph{\textbf{Multi-Label Recognition with Partial Labels}} For MLR-PL task, the training set contains three label states where $y_i^c$ is assigned to 1 if label $c$ exists in the $i$-th image, assigned to -1 if it does not exist, and assigned to 0 if it is unknown/missing. Our goal in this task is to train a robust multi-label classification model given partial labels, and further assign the correct label for a test image. Since the BCE loss as Eq. (\ref{loss}) is dependent on the number of categories and does not apply to partially labeled data, we follow previous work \cite{durand2019learning} to employ the partial BCE loss as our objective function. Specially, given the predicted probability distribution $\textbf{p}_i=[p_i^1,p_i^2,...,p_i^C]$ of the image $i$ and its ground truth $\textbf{y}_i$, the object function can be defined as
\begin{gather}
 \mathcal{L}(\textbf{p}_i, \textbf{y}_i)=\frac{g(r_{\textbf{y}_i})}{C} \sum_{c=1}^{C}[\mathbbm{1}_{[y_i^c=1]}log(p_i^c)+\mathbbm{1}_{[y_i^c=-1]}log(1-p_i^c)]\nonumber\\
  g(r_{\textbf{y}_i})=\alpha{r_{\textbf{y}_i}^{\mu}}+\theta,
 \label{partial BCE}
 \end{gather} 
where $\mathbbm{1}[\cdot]$ is an indicator function whose value is 1 if the argument is positive and is 0 otherwise. $r_{\textbf{y}_i} \in [0,1]$ is the proportion of known labels in $\textbf{y}_i$. And $g(\cdot)$ is a normalization function with respect to the label proportion, where $\alpha$, $\theta$ and $\mu$ are the hyper-parameters. It is worth noting that by learning category co-occurrence and modeling instance-label correspondences, our framework can not only transfer semantic knowledge of the given labels to missing labels within each image, but also complement missing labels across different images. 

\paragraph{\textbf{Multi-Label Few-Shot Learning}} For ML-FSL task, the dataset contains a set of $C_b$ base categories with sufficient training samples and a set of $C_n$ novel categories with limited training samples, \textit{e.g.}, 1, 2 or 5.  
Since the training samples for the base and novel categories are extremely unbalanced, it will inevitably lead to poor performances if the cross-entropy loss is directly used to guide the training of all samples. Thus, we follow the previous work \cite{alfassy2019laso,chen2020knowledge} to adopt a two-stage process to train the proposed framework.

In the first stage, we train the framework using the training samples of the base categories, which has the same training protocol as generic MLIR. Specifically, given a sample $i$ from the $C_b$ categories, we can obtain 
the corresponding prediction score $p_i^c(c=1,2,...,C_b)$ for the $c$-th categories. Then, we define the loss similar to Eq. (\ref{loss}), expressed as   
\begin{gather}
 \mathcal{L}=\sum_{i=1}^{N_b}\sum_{c=1}^{C_b}w^c[y_i^c\log(p_i^c)+(1-y_i^c)\log(1-p_i^c))]\nonumber\\
 w^c=y_i^c\cdot e^{\beta(1-r^c)}+(1-y_i^c)\cdot e^{\beta r^c},
 \end{gather}  
where $N_b$ is the number of training samples from the base set. 

In the second stage, we fix the parameters of the feature extractor backbone and train the CSAC and ILMS modules using the novel set. Notably, we introduce an asymmetric focal loss \cite{ridnik2021asymmetric} for multi-label few-shot recognition in this stage, which explicitly solves the positive-negative imbalance problem:      
\begin{equation}
\left\{
\begin{aligned}
\mathcal{L}_{+,i,c}=(1-p_{i,c})^{\gamma_+}log(p_{i,c}), \\
\mathcal{L}_{-,i,c}=(\hat{p}_{i,c})^{\gamma_{-}}log(1-\hat{p}_{i,c}),
\end{aligned}
\right.
\label{focal loss}
\end{equation}
\begin{equation}
\hat{p}_{i,c}=max(p_{i,c}-m, 0),
\label{margin}
\end{equation}
where $p_{i,c}(c=1,2,...,C_n)$ is the prediction score for the $c$-th category of the sample $i$ from the $C_n$ categories. $\gamma_+$ and $\gamma_{-}$ are respectively the positive and negative focusing parameters. $m$ is a tunable hyper-parameter performing hard threshold for very easy negative instances.

Since we focus on emphasizing the contribution of positive instances, we commonly set $\gamma_{-} \textgreater \gamma_{+}$. Asymmetric focusing decouples the decay rates of positive and negative instances. By means of this way, our model can better control the contribution of positive and negative instances, and aid the network learn significant features from positive instances, despite their rarity.
\begin{equation}
\mathcal{L}=\sum_{i=1}^{N_n}\sum_{c=1}^{C_n}y_{i,c}\mathcal{L}_{+,i,c}+(1-y_{i,c})\mathcal{L}_{-,i,c},
\end{equation}
where $N_n$ is the number of training samples from the novel set. In summary, by extracting semantic features and exploiting the correlation between novel categories and base categories, our proposed model can classify multiple labels only using a few samples effectively.  

\section{Experiments}
In this section, we first conduct extensive experiments on various benchmarks to evaluate the performance of the proposed framework over existing state-of-the-art methods. Then, we perform ablative studies to further discuss and analyze the contribution of the crucial components in our ML-SGM.

\subsection{Datasets}

\paragraph{\textbf{PASCAL VOC 2007 Benchmark}} PASCAL Visual Object Classes Challenge 2007 (VOC 2007) \cite{everingham2010pascal} is the most widely used benchmark for multi-label image recognition. It contains 9,963 images from 20 object categories, which is divided into \textit{train}, \textit{val} and \textit{test} sets. We follow the common setting \cite{chen2019multi} to train our network on the \textit{train-val} sets, and evaluate its classification performance on the \textit{test} set. To compare with other state-of-the-art approaches, we calculate the results of average precision (AP) and mean average precision (mAP).

\paragraph{\textbf{MS-COCO Benchmark}} Microsoft COCO \cite{lin2014microsoft} is initially constructed for object detection and segmentation, and has recently been employed to evaluate multi-label image recognition task in terms of scene comprehension. The training set consists of 82,783 images, including common objects in the scenes. The objects are divided into 80 categories, and each image has about 2.9 labels. Because the ground-truth labels of test set are not available, we assess all approaches on the validation set (40,504 images). 

\paragraph{\textbf{NUS-WIDE Benchmark}} The NUS-WIDE \cite{chua2009nus} is a web dataset with 269,648 images and 5018 labels. There are 1000 classes remaining after removing the noise and the rare labels. These images are further manually labeled into 81 concepts with an average of 2.4 concepts per image. According to its standard settings, we employ 161,789 images to train and 107,859 images to test. 

\paragraph{\textbf{Visual Genome Benchmark}} The Visual Genome \cite{krishna2017visual} is a dataset that contains 108,249 images and covers 80,138 categories. Since most categories have very few samples, we merely consider the 200 most frequent categories to obtain a VG-200 subset for MLIR-PL task. Meanwhile, we also follow previous work \cite{chen2020knowledge} to consider the 500 most frequent categories, resulting in a VG-500 subset for ML-FSL task. Moreover, because there is no official train/-val split, we randomly select 10,000 images as the test set and the remaining 98,249 images are used as the training set. 

\begin{table*}[htbp]
\caption{Comparisons of AP and mAP on the VOC 2007 dataset. The performance of our method is based on a report with a resolution of $448 \times 448$. Among the different metrics, the \textcolor{red}{red} numbers represent the best results, and the \textcolor{blue}{blue} numbers represent the sub-optimal results. Best viewed in color. * denotes the performance of our implementation.}
\begin{center}
\setlength{\tabcolsep}{0.5mm}{
\begin{tabular}{c|c|c|c|c|c|c|c|c|c|c|c|c|c|c|c|c|c|c|c|c|c}
\toprule[1pt]
\textbf{Methods} &\emph{\textbf{aero}} &\emph{\textbf{bike}} &\emph{\textbf{bird}} &\emph{\textbf{boat}} &\emph{\textbf{bottle}} &\emph{\textbf{bus}} &\emph{\textbf{car}} &\emph{\textbf{cat}} &\emph{\textbf{chair}} &\emph{\textbf{cow}} &\emph{\textbf{table}} &\emph{\textbf{dog}} &\emph{\textbf{horse}} &\emph{\textbf{motor}} &\emph{\textbf{person}} &\emph{\textbf{plant}} &\emph{\textbf{sheep}} &\emph{\textbf{sofa}} &\emph{\textbf{train}} &\emph{\textbf{tv}} &\textbf{mAP} \\\hline\hline
HCP \cite{wei2015hcp} &98.6 &97.1 &98.0 &95.6 &75.3 &94.7 &95.8 &97.3 &73.1 &90.2 &80.0 &97.3 &96.1 &94.9 &96.3 &78.3 &94.7 &76.2 &97.9 &91.5 &90.9 \\
CNN-RNN \cite{wang2016cnn} &96.7 &83.1 &94.2 &92.8 &61.2 &82.1 &89.1 &94.2 &64.2 &83.6 &70.0 &92.4 &91.7 &84.2 &93.7 &59.8 &93.2 &75.3 &\textcolor{red}{99.7} &78.6 &84.0 \\
ResNet-101* \cite{he2016deep} &99.5 &97.7 &97.8 &96.4 &65.7 &91.8 &96.1 &97.6 &74.2 &80.9 &85.0 &98.4 &96.5 &95.9 &98.4 &70.1 &88.3 &80.2 &98.9 &89.2 &89.9 \\
RNN-Attention \cite{wang2017multi} &98.6 &97.4 &96.3 &96.2 &75.2 &92.4 &96.5 &97.1 &76.5 &92.0 &87.7 &96.8 &97.5 &93.8 &98.5 &81.6 &93.7 &82.8 &98.6 &89.3 &91.9 \\
ML-GCN \cite{chen2019multi} &99.6 &98.3 &97.9 &97.6 &78.2 &92.3 &97.4 &97.4 &79.2 &94.4 &86.5 &97.4 &97.9 &\textcolor{blue}{97.1} &98.7 &84.6 &95.3 &83.0 &98.6 &90.4 &93.1 \\
SSGRL \cite{chen2019learning} &99.5 &97.1 &97.6 &97.8 &82.6 &94.8 &96.7 &98.1 &78.0 &\textcolor{blue}{97.0} &85.6 &97.8 &\textcolor{blue}{98.3} &96.4 &98.8 &84.9 &96.5 &79.8 &98.4 &92.8 &93.4 \\
TSGCN \cite{xu2020joint} &98.9 &\textcolor{blue}{98.5} &96.8 &97.3 &\textcolor{red}{87.5} &94.2 &97.4 &97.7 &\textcolor{blue}{84.1} &92.6 &\textcolor{red}{89.3} &98.4 &98.0 &96.1 &98.7 &84.9 &96.6 &\textcolor{red}{87.2} &98.4 &93.7 &94.3 \\
ADD-GCN* \cite{ye2020attention} &99.7 &98.5 &97.6 &\textcolor{blue}{98.4} &80.6 &94.1 &96.6 &98.1 &80.4 &94.9 &85.7 &97.9 &97.9 &96.4 &99.0 &80.2 &\textcolor{blue}{97.3} &85.3 &98.9 &94.1 &93.6\\ 
DSDL \cite{zhou2021deep} &\textcolor{blue}{99.8} &\textcolor{blue}{98.7} &98.4 &97.9 &81.9 &\textcolor{blue}{95.4} &97.6 &98.3 &83.3 &95.0 &\textcolor{blue}{88.6} &98.0 &97.9 &95.8 &99.0 &86.6 &95.9 &\textcolor{blue}{86.4} &98.6 &94.4 &94.4\\ 
SST \cite{chen2022sst} &\textcolor{blue}{99.8} &98.6 &\textcolor{red}{98.9} &\textcolor{blue}{98.4} &85.5 &94.7 &\textcolor{blue}{97.9} &\textcolor{blue}{98.6} &83.0 &96.8 &85.7 &\textcolor{blue}{98.8} &\textcolor{red}{98.9} &95.7 &\textcolor{blue}{99.1} &\textcolor{blue}{85.4} &96.2 &84.3 &\textcolor{blue}{99.1} &\textcolor{blue}{95.0} &\textcolor{blue}{94.5}\\ \hline\hline
Ours &\textcolor{red}{99.9} &\textcolor{red}{98.8} &\textcolor{blue}{98.5} &\textcolor{red}{98.6} &\textcolor{blue}{86.3} &\textcolor{red}{96.0} &\textcolor{red}{98.0} &\textcolor{red}{99.2} &\textcolor{red}{84.5} &\textcolor{red}{97.6} &87.7 &\textcolor{red}{99.2} &\textcolor{red}{98.9} &\textcolor{red}{97.2} &\textcolor{red}{99.3} &\textcolor{red}{86.0} &\textcolor{red}{98.3} &\textcolor{red}{87.2} &\textcolor{blue}{99.1} &\textcolor{red}{95.3} &\textcolor{red}{95.2}\\ 
\bottomrule[1pt]
\end{tabular}}
\label{voc2007}
\end{center}
\vspace{-3mm}
\end{table*}

\begin{table*}[htbp]
\caption{Comparisons of AP and mAP on the MS-COCO dataset. $R_{train}$ and $R_{test}$ denote resolution used in training and testing stage. The \textcolor{red}{red} numbers represent the best results, and the \textcolor{blue}{blue} numbers represent the sub-optimal results. Best viewed in color.  * denotes the performance of our implementation. '-' indicates the results are not reported in the original literature.}
\begin{center}                                                     
\setlength{\tabcolsep}{1.8mm}{
\begin{tabular}{c|c|c|c|c|c|c|c|c|c|c|c|c|c|c}
\toprule[1pt]
\multirow{2}{*}{\textbf{Methods}}  &\multirow{2}{*}{($\textbf{R}_{\textbf{train}}$, $\textbf{R}_{\textbf{test}}$)} & \multirow{2}{*}{\textbf{mAP}} &\multicolumn{6}{c|}{\textbf{All}} &\multicolumn{6}{c}{\textbf{Top-3}}  \\\cline{4-15}
  & & & \textbf{CP} & \textbf{CR} & \textbf{CF1} & \textbf{OP} & \textbf{OR} & \textbf{OF1} & \textbf{CP} & \textbf{CR} & \textbf{CF1} & \textbf{OP} & \textbf{OR} & \textbf{OF1}  \\\hline\hline
CNN-RNN \cite{wang2016cnn} &(—, —) &61.2 &- &- &- &- &- &- &66.0 &55.6 &60.4 &69.2 &66.4 &67.8 \\
Order-Free RNN \cite{chen2018order} &(—, —) &- &- &- &- &- &- &- &71.6 &54.8 &62.1 &74.2 &62.2 &67.7 \\
ResNet-101* \cite{he2016deep} &(448, 448) &77.3 &80.2 &66.7 &72.8 &83.9 &70.8 &76.8 &84.1 &59.4 &69.7 &89.1 &62.8 &73.6 \\ 
SRN \cite{zhu2017learning} &(224, 224) &77.1 &81.6 &65.4 &71.2 &82.7 &69.9 &75.8 &85.2 &58.8 &67.4 &87.4 &62.5 &72.9 \\
ML-GCN \cite{chen2019multi} &(448, 448) &83.0 &85.1 &72.0 &78.0 &85.8 &75.4 &80.3 &89.2 &64.1 &74.6 &90.5 &66.5 &76.7 \\
CMA \cite{you2020cross} &(448, 448) &83.4 &82.1 &73.1 &77.3 &83.7 &76.3 &79.9 &87.2 &64.6 &74.2 &89.1 &66.7 &76.3 \\
TSGCN \cite{xu2020joint} &(448, 448) &83.5 &81.5 &72.3 &76.7 &84.9 &75.3 &79.8 &84.1 &67.1 &74.6 &89.5 &69.3 &78.1 \\
DSDL \cite{zhou2021deep} &(448, 448) &81.7 &88.1 &62.9 &73.4 &89.6 &65.3 &75.6 &84.1 &\textcolor{red}{70.4} &76.7 &85.1 &\textcolor{red}{73.9} &79.1 \\
SST \cite{chen2022sst} &(448, 448) &84.2 &86.1 &72.1 &78.5 &87.2 &75.4 &80.8 &89.8 &64.1 &74.8 &91.5 &66.4 &76.9 \\
Ours &(448, 448) &85.1 &87.2 &74.2 &80.2 &88.7 &76.5 &82.1 &90.1 &67.5 &77.2 &92.3 &71.2 &80.4\\\hline\hline
ADD-GCN \cite{ye2020attention} &(448, 576) &85.2 &84.7 &\textcolor{red}{75.9} &80.1 &84.9 &\textcolor{red}{79.4} &82.0 &88.8 &66.2 &75.8 &90.3 &68.5 &77.9\\
Ours &(448, 576) &\textcolor{blue}{86.0} &88.1 &75.0 &\textcolor{blue}{81.0} &89.4 &\textcolor{blue}{77.2} &\textcolor{blue}{82.9} &91.0 &68.6 &\textcolor{blue}{78.2} &93.0 &72.4 &\textcolor{blue}{81.4}\\\hline\hline
SSGRL \cite{chen2019learning} &(576, 576) &83.6 &\textcolor{red}{89.5} &68.3 &76.9 &\textcolor{red}{91.2} &70.7 &79.3 &\textcolor{red}{91.9} &62.1 &73.0 &\textcolor{blue}{93.6} &64.2 &76.0 \\
C-Tran \cite{lanchantin2021general} &(576, 576) &85.1 &86.3 &74.3 &79.9 &87.7 &76.5 &81.7 &90.1 &65.7 &76.0 &92.1 &71.4 &77.6 \\
Ours &(576, 576) &\textcolor{red}{86.3} &\textcolor{blue}{88.4} &\textcolor{blue}{75.6} &\textcolor{red}{81.5} &\textcolor{blue}{89.8} &\textcolor{blue}{77.2} &\textcolor{red}{83.0} &\textcolor{blue}{91.4} &\textcolor{blue}{68.9} &\textcolor{red}{78.6} &\textcolor{red}{93.8} &\textcolor{blue}{72.5} &\textcolor{red}{81.8}\\
\bottomrule[1pt]
\end{tabular}}
\label{coco}
\end{center}
\vspace{-3mm}
\end{table*}

\subsection{Evaluation Metrics}
Following the traditional setting \cite{chen2019multi,chen2019learning}, we employ six widely used multi-label metrics to evaluate our method and each comparing method. Specifically, we compute the average per-class precision (CP), recall (CR), F1 (CF1) and the average overall precision (OP), recall (OR), F1 (OF1) with the assumption that a projected label is positive if the estimated probability is larger than 0.5. We also present the results of top-3 labels to fairly compare with state-of-the-art approaches.  
Furthermore, we calculate and show the average precision (AP) and mean average precision (mAP) used to evaluate the classification accuracy of multi-label images. 

\subsection{Implementation Details} 
In the CSAC module, we employ ResNet-101 \cite{he2016deep} as the backbone network to extract visual features per image. To save computational cost, we further convert the number of channels of feature map $\textbf{F}$ to 1024 via a convolution layer. For label initial embeddings, we adopt 300-dim GloVe \cite{pennington2014glove} trained on the Wikipedia dataset. In the ILMS module, the Graph Convolution Layer consists of $k$ convolution modules, where they are stacked to aggregate the information of $k^{th}$-order neighborhoods. In our experiments, $k$ is 2 and the output dimension of the corresponding convolution module is 512 and 256, respectively. During the training stage, the input images are adjusted to 640×640, and randomly
cropped into 448× 448 with random horizontal flips for data augmentation. Instead, testing images are center cropped. All modules of our framework are implemented in PyTorch 6.0 and the optimizer is SGD with momentum 0.9. Weight decay is 10$^{-4}$. The initial learning rate is 0.01, which decays by a factor of 10 for every 30 epochs. The hyperparameter $\beta$ in the Eq. (\ref{loss}) is set to 0 in VOC 2007 dataset and 0.4 in MS-COCO, NUS-WIDE and Visual Genome 500 datasets. The hyperparameter $\alpha$, $\theta$ and $\mu$ in Eq. (\ref{partial BCE}) are set to -4.45, 5.45 and 1 on VOC 2007, MS-COCO and Visual Genome 200 datasets for MLR-PL task. And the hyperparameter $\gamma_+$, $\gamma_{-}$ and $m$ in Eq. (\ref{focal loss}) and (\ref{margin}) are set to 0, 4 and 0.05 on both MS-COCO and Visual Genome 500 datasets for ML-FSL task. Unless otherwise stated, the choice of these hyperparameters is carefully turned with the validation set in target datasets.

\begin{figure*}[!ht]
\centering
\setlength{\tabcolsep}{0.4mm}{
\begin{tabular}{|c:c:c|c:c:c|}
\hline
 & & & & &\\
\includegraphics[width=28mm,height=18mm]{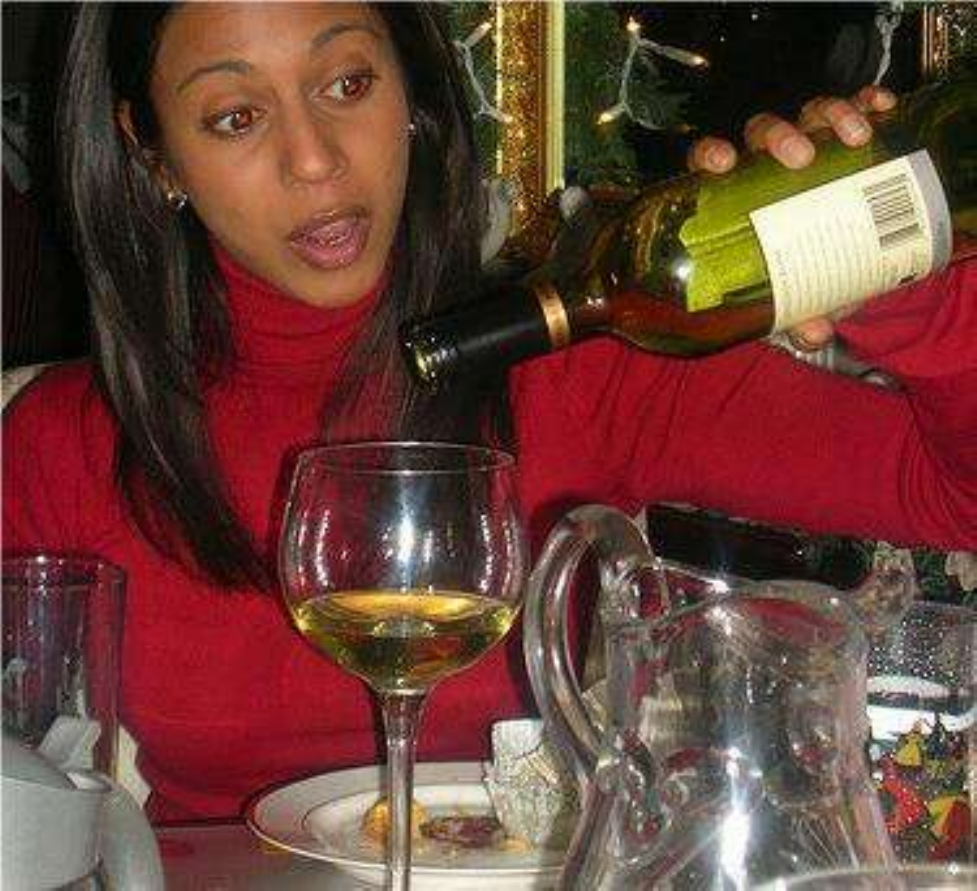}
&\includegraphics[width=28mm,height=18mm]{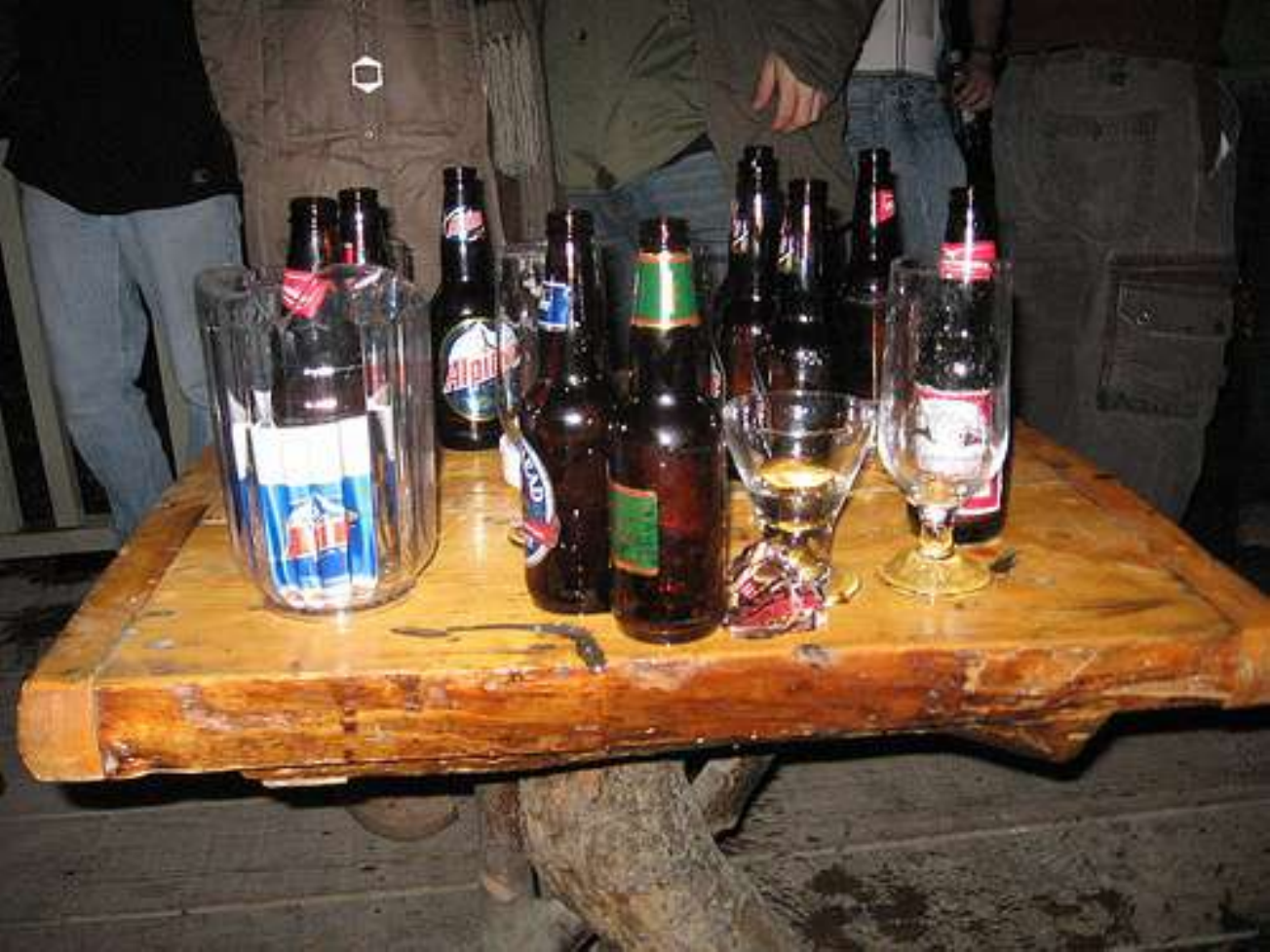}
&\includegraphics[width=28mm,height=18mm]{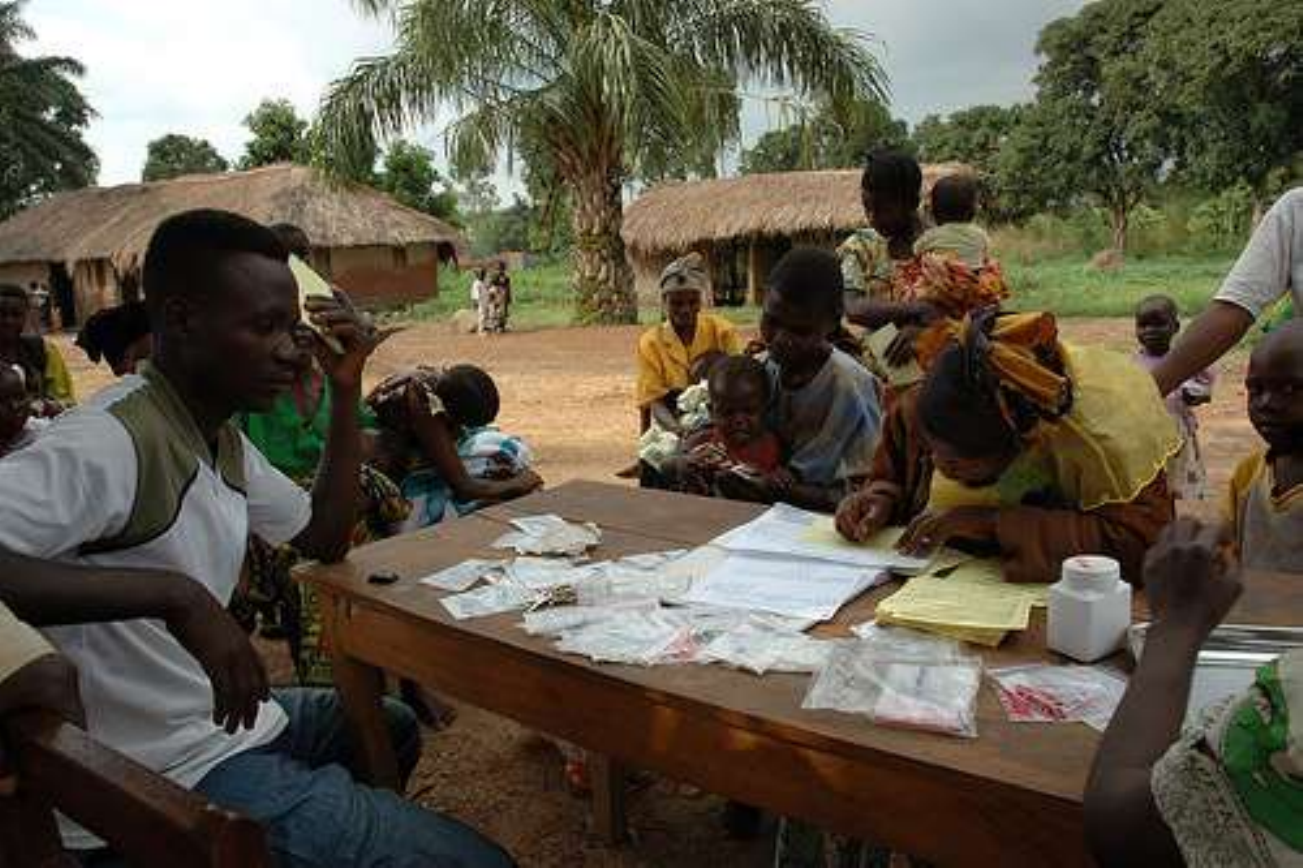}
&\includegraphics[width=28mm,height=18mm]{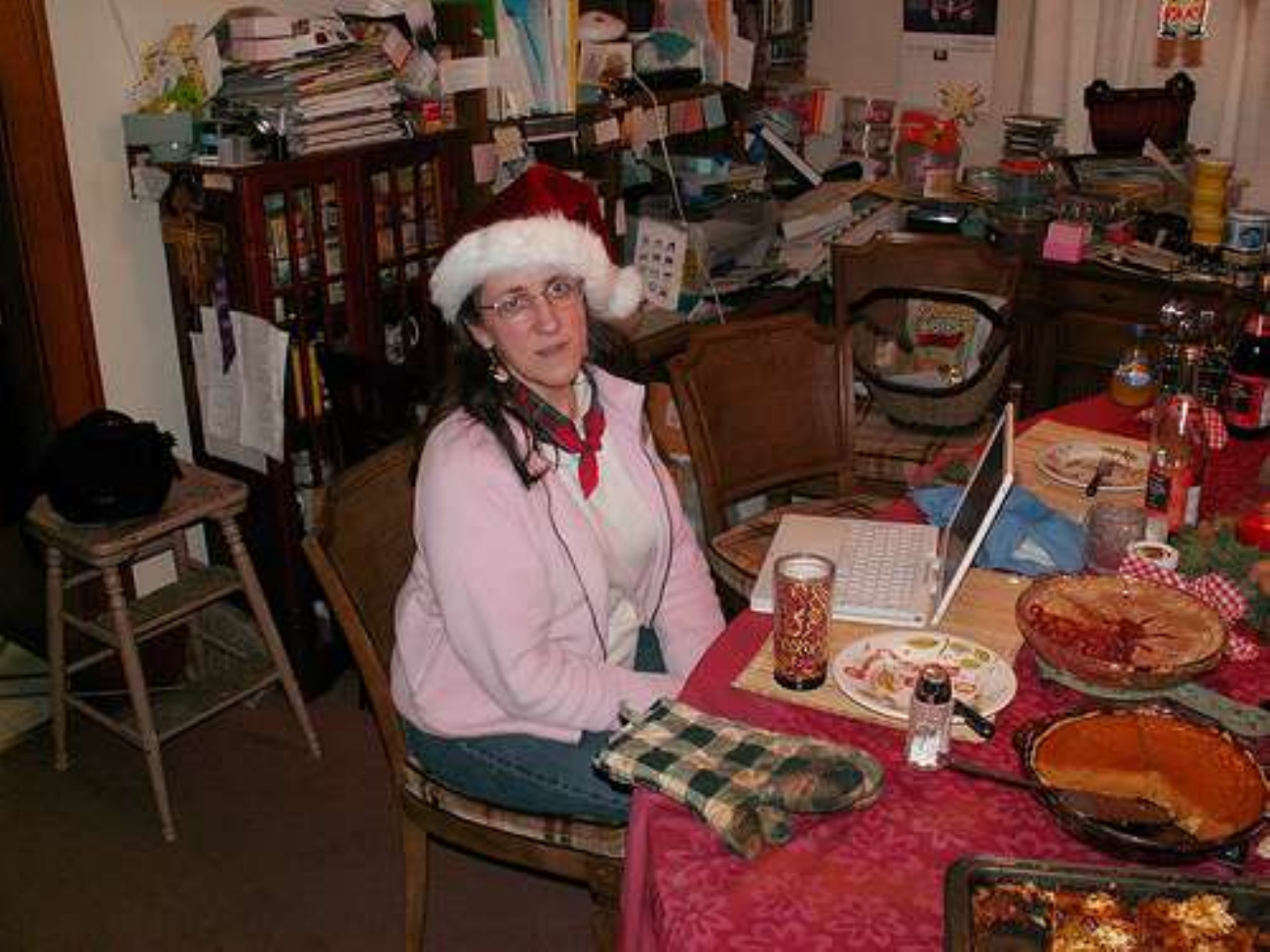}
&\includegraphics[width=28mm,height=18mm]{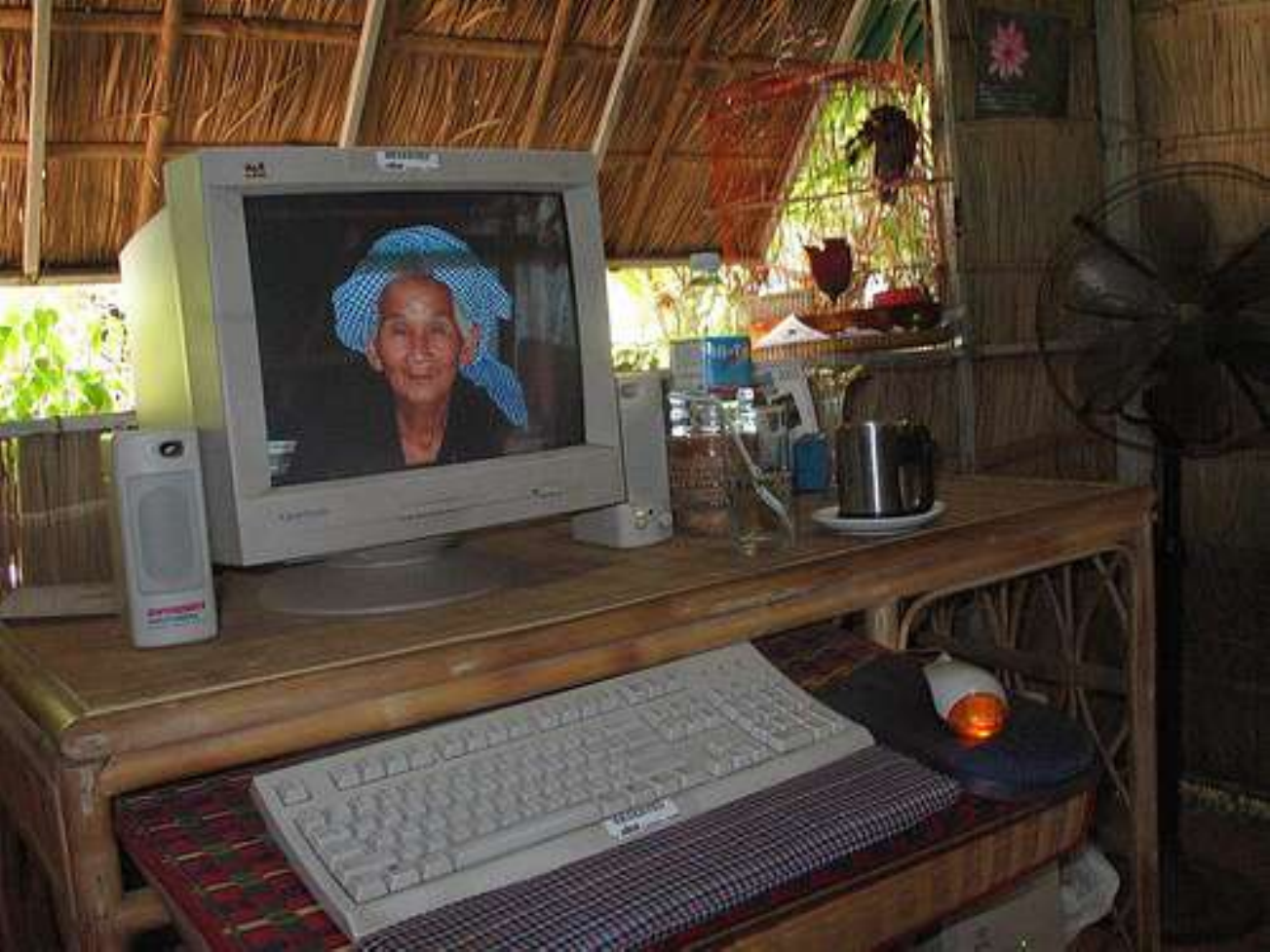}
&\includegraphics[width=28mm,height=18mm]{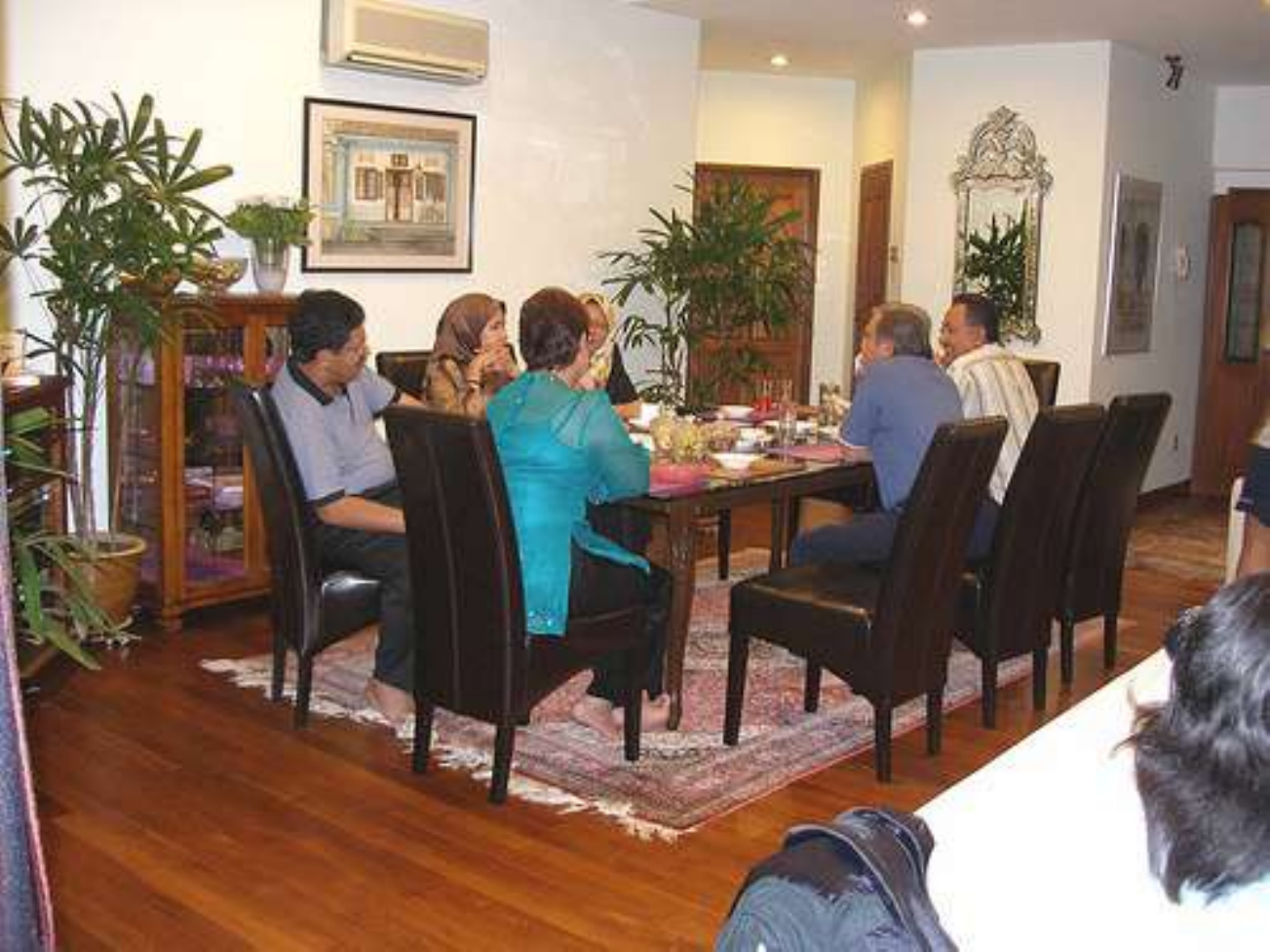}\\
\scriptsize{person, bottle, \textcolor{red}{dining table}}
&\scriptsize{person, bottle, \textcolor{red}{dining table}}
&\scriptsize{person, \textcolor{red}{dining table}}
&\scriptsize{person, dining table, \textcolor{red}{chair}, \textcolor{green}{bottle}}
&\scriptsize{person, tvmonitor, \textcolor{green}{bottle}}
&\scriptsize{person, chair, plant, table,  \textcolor{green}{bottle}} \\
\textcolor{blue}{\scriptsize{person, bottle}}
&\textcolor{blue}{\scriptsize{person, bottle}}
&\textcolor{blue}{\scriptsize{person}}
&\textcolor{blue}{\scriptsize{bottle, person, dining table}}
&\textcolor{blue}{\scriptsize{tvmonitor, person, bottle}}
&\textcolor{blue}{\scriptsize{chair, table, person, plant, bottle}} \\ \hline
\end{tabular}}
\caption{Some incorrect classification results on the \textit{dinner table} and \textit{bottle} categories. The last row represents the ground truth label of the corresponding image. The second row is the predicted labels of our model. The \textcolor{red}{red} indicates false positives and the \textcolor{green}{green} indicates false negatives. Best viewed in color and zoom in.}
\label{incorrect_visual}
\end{figure*}

\subsection{Results on Multi-Label Image Recognition}

\paragraph{\textbf{Comparison on Pascal VOC 2007}} 
We first report the AP of each label and mAP over all labels in Table \ref{voc2007}. 
Despite the fact that the VOC 2007 dataset is less complicated and relatively small size, our method still 
sets the new SOTA in terms of mAP and significantly outperforms the second best method SST \cite{chen2022sst} by 0.7\% (note, SST only outperforms DSDL \cite{zhou2021deep} by 0.1\%). And the proposed method achieves the best results on 16 out of 20 categories. In Particular, it achieves meaningful improvements in several difficult categories, such as \textit{bird}, \textit{chair}, \textit{plant} and \textit{sofa} with 98.5\%, 84.5\%, 86.0\% and 87.2\% in terms of AP. This phenomenon illustrates that exploiting the co-occurrence and matching relationships among instances and labels with graph neural network is effective for MLIR task. Note that * is employed to denote the performance result of our implementation based on author-supplied code. For fair comparisons, we adopt the same ResNet-101 network as the backbone and follow exactly the same train/val split settings. In addition, the performance results of other comparison methods are directly taken from the source publications.

Here Table \ref{voc2007} shows that our proposed method does not perform well on the category \textit{bottle} and \textit{dinner table} compared to TSGCN \cite{xu2020joint}. The main reason is due to the higher model complexity and computational budget required for TSGCN. For example, TSGCN uses a pre-trained ResNet-101 as a feature extractor in the input space and utilizes another pre-trained Mask R-CNN in the output space. However, our method follows an easier one-stage framework as in \cite{chen2022sst,chen2019multi,zhou2021deep,ye2020attention,xu2020joint,chen2019learning} which only uses one pre-trained ResNet-101. Despite the higher learning capacity of TSGCN, our method still outperforms it by a large margin in terms of mAP.

On the other hand, we visualize some incorrect classification results on the \textit{dinner table} and \textit{bottle} categories in Figure \ref{incorrect_visual}. The first three examples of Figure \ref{incorrect_visual} show that the \textit{dinner
table} is incorrectly predicted as positive. The reason for this phenomenon can be attributed to two aspects: (i) some images in the dataset mislead our model due to the missing annotation of “\textit{dinner table}”; (ii) similar categories share similar characteristics and are easily confused by the model, e.g.  “\textit{dinner table}” and “\textit{table}”. In addition, it can be seen that the \textit{bottle} category presents a higher false negative from the last three examples in Figure \ref{incorrect_visual}. Further analysis reveals that \textit{bottle} category is often filtered out when we select class activation maps or instance regions with prediction scores greater than the threshold $\gamma$. Since there are fewer pixels representing small objects, they will disappear in the process of downsampling the feature map and lead to low prediction scores.

\begin{table}[t]
\caption{Comparisons with state-of-the-art methods on the NUS-WIDE dataset. * denotes the performance of our implementation. '-' indicates the results are not reported in the original literature.}
\begin{center}
\setlength{\tabcolsep}{2.2mm}{
\begin{tabular}{c|c|c|c|c|c}
\toprule[1pt]
\multirow{2}{*}{\textbf{Methods}} &\multicolumn{3}{c|}{\textbf{All}} &\multicolumn{2}{c}{\textbf{Top-3}} \\ \cline{2-6}
               & \textbf{mAP} & \textbf{CF1} & \textbf{OF1} &\textbf{CF1} &\textbf{OF1}  \\\hline

CNN-RNN \cite{wang2016cnn} &56.1 &- &- &34.7 &55.2  \\
SRN \cite{zhu2017learning} &61.8 &56.9 &73.2 &47.7 &62.2  \\
MLIC-KD-WSD \cite{liu2018multi} &60.1 &58.7 &\textcolor{blue}{73.7} &53.8 &71.1  \\
ML-ZSL \cite{lee2018multi} &- &- &- &45.7 &- \\
PLA \cite{yazici2020orderless} &- &56.2 &- &- &\textcolor{red}{72.3} \\ 
ResNet-101* \cite{he2016deep} &59.8 &55.7 &72.5 &48.9 &62.2\\
CMA \cite{you2020cross} &61.4 &\textcolor{blue}{60.5} &\textcolor{red}{73.8} &55.7 &69.5 \\
SST \cite{chen2022sst} &\textcolor{blue}{63.5} &59.6 &73.2 &\textcolor{blue}{55.9} &68.8 \\\hline
Ours &\textcolor{red}{64.6} &\textcolor{red}{62.4} &72.5 &\textcolor{red}{57.3} &\textcolor{blue}{71.7} \\
\bottomrule[1pt]
\end{tabular}}
\label{nus-wide}
\end{center}
\vspace{-3mm}
\end{table}

\begin{figure*}[!ht]
\centering
\setlength{\tabcolsep}{0.4mm}{
\begin{tabular}{|c:c:c:c:c:c|}
\hline
 & & & & &\\
\includegraphics[width=28mm,height=18mm]{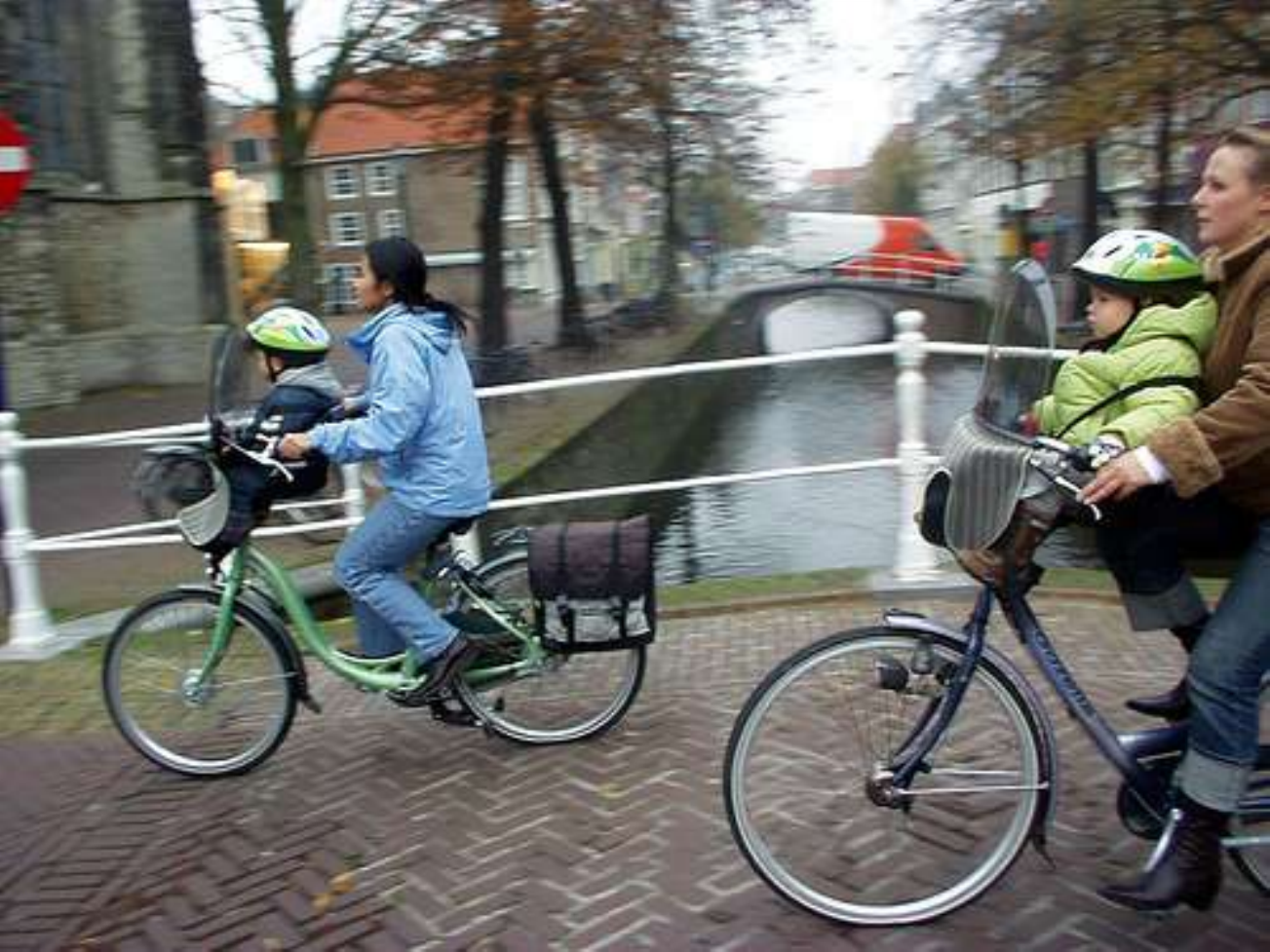}
&\includegraphics[width=28mm,height=18mm]{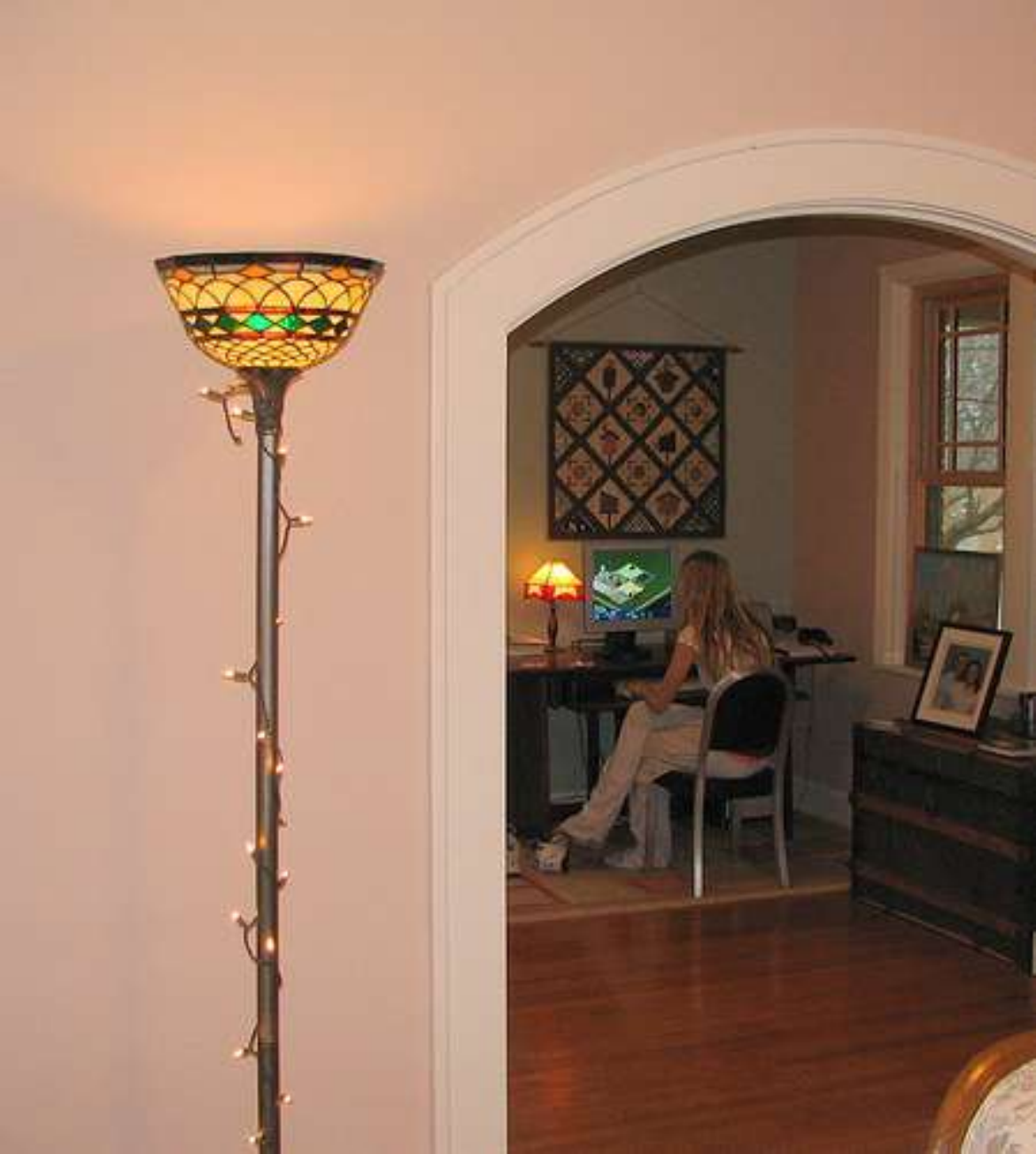}
&\includegraphics[width=28mm,height=18mm]{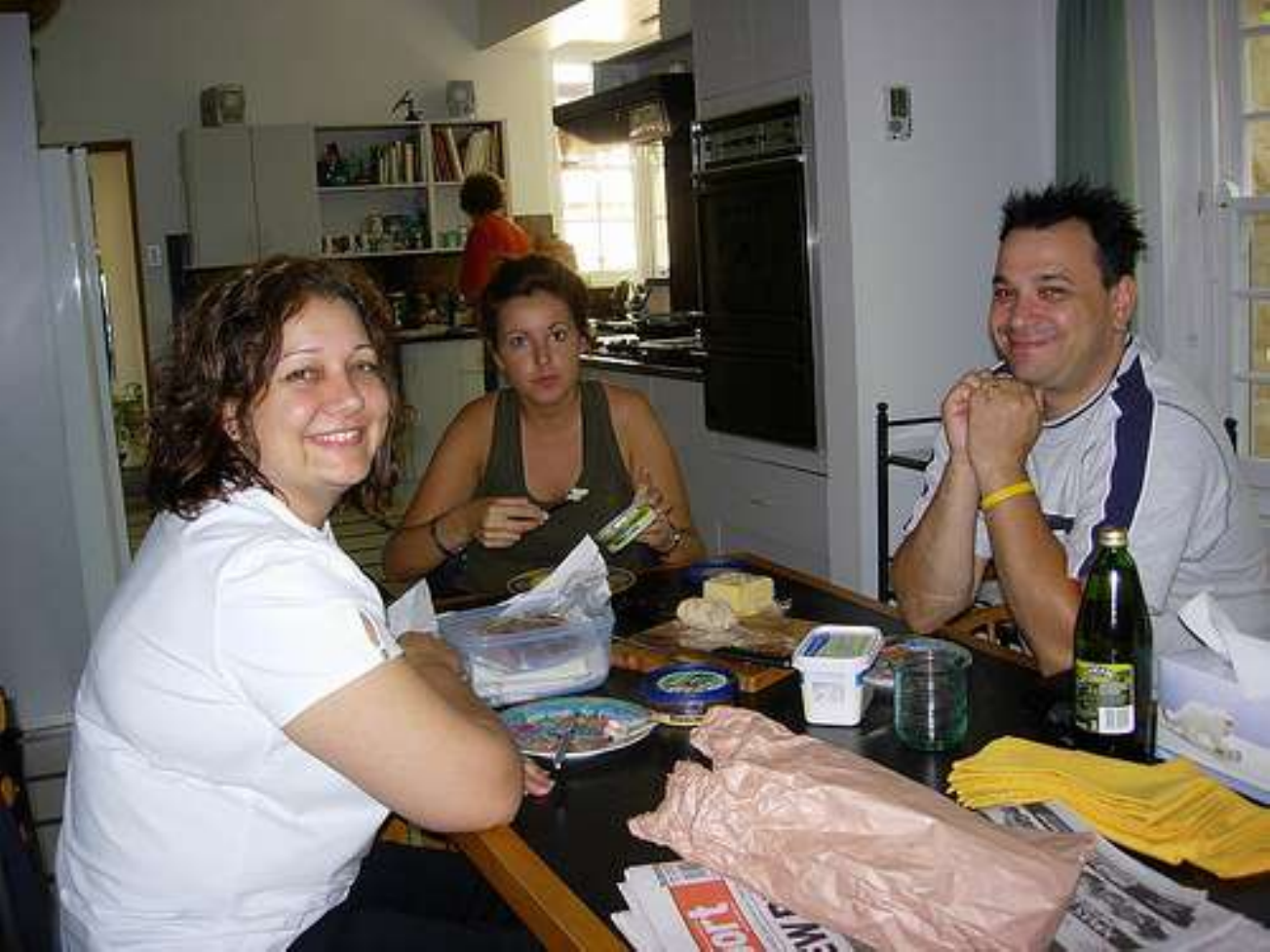}
&\includegraphics[width=28mm,height=18mm]{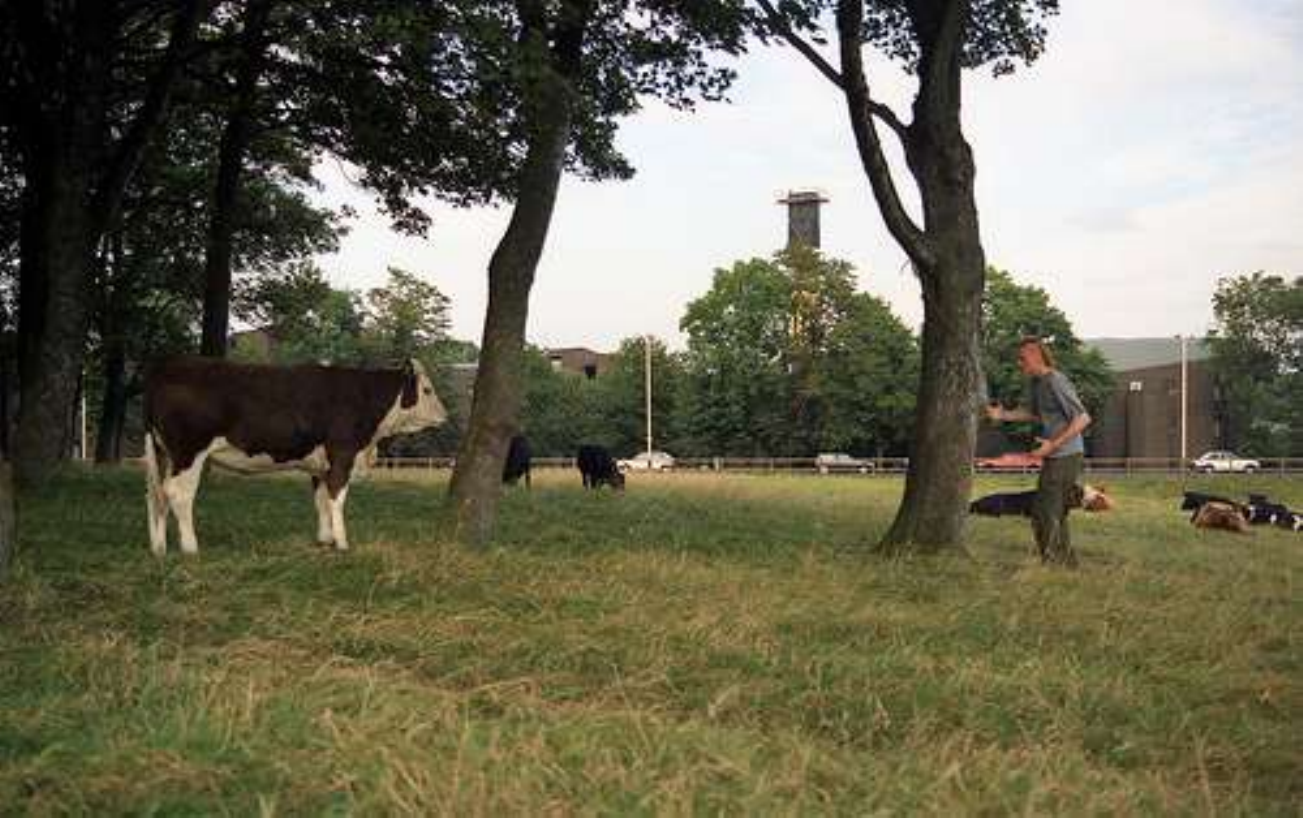}
&\includegraphics[width=28mm,height=18mm]{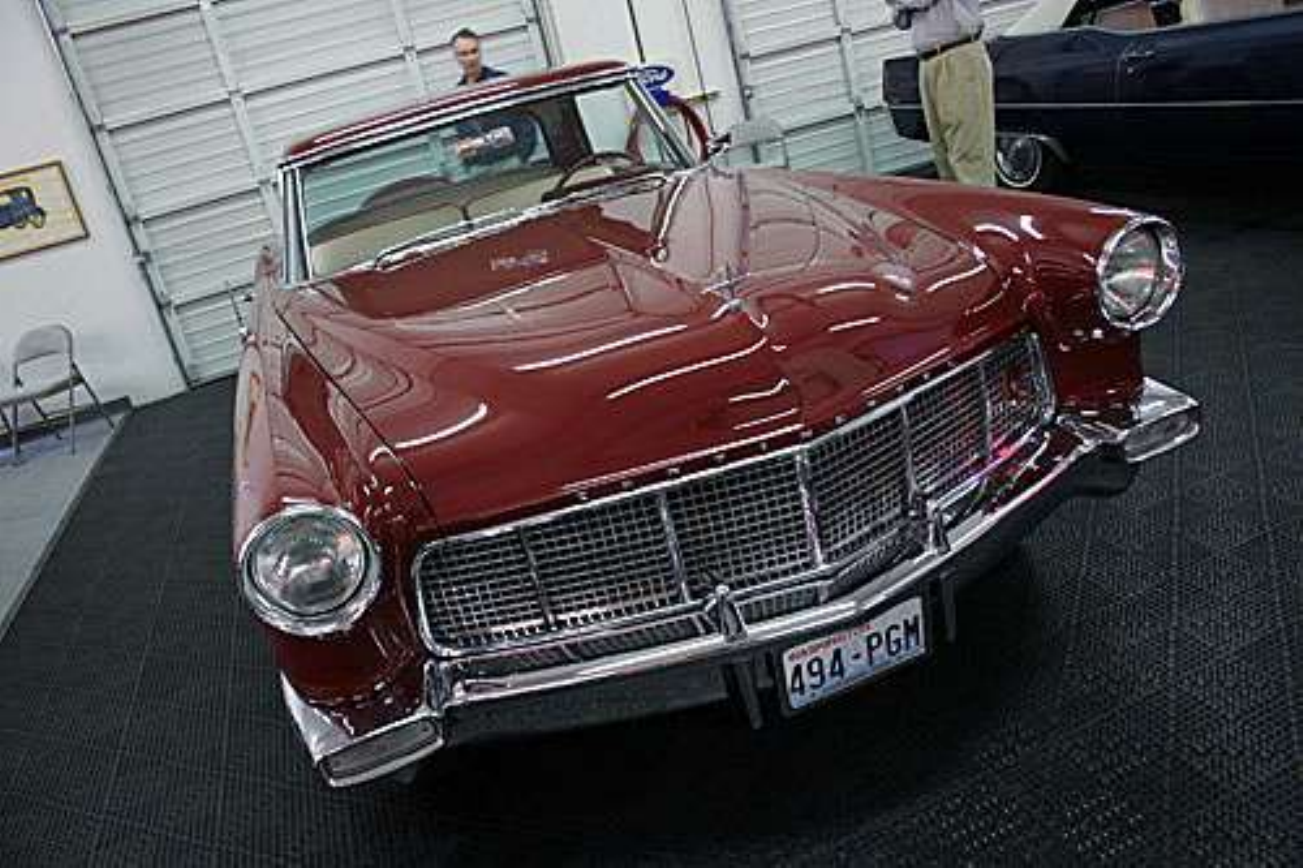}
&\includegraphics[width=28mm,height=18mm]{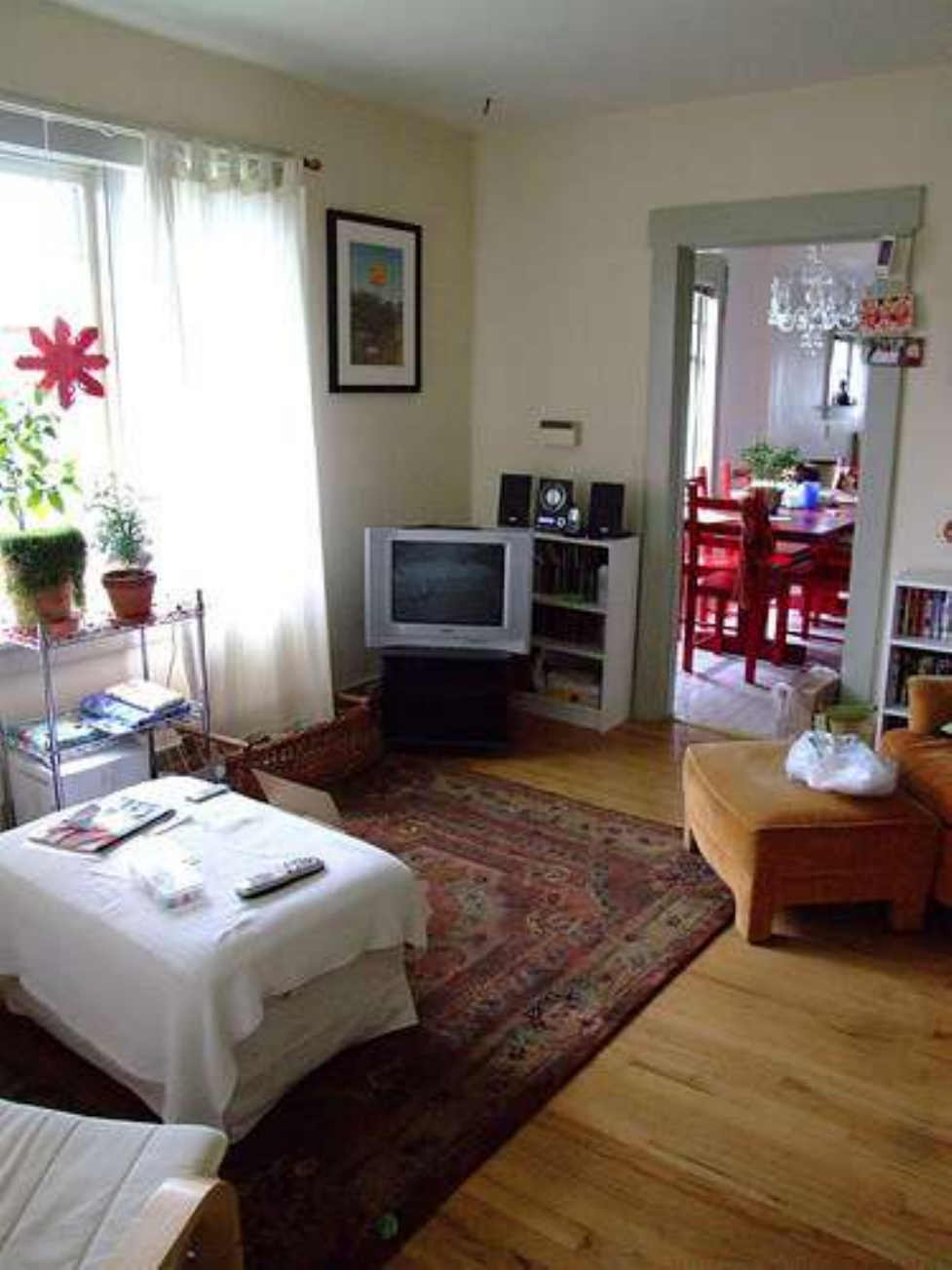}\\
\textcolor{red}{\scriptsize{person,bicycle,car}}
&\textcolor{red}{\scriptsize{person,chair,tvmonitor}}
&\textcolor{red}{\scriptsize{bottle,person,dining table}}
&\textcolor{red}{\scriptsize{cow,person,car}}
&\textcolor{red}{\scriptsize{car,person,chair}}
&\textcolor{red}{\scriptsize{tvmonitor,chair,potted plant}} \\
\textcolor{blue}{\scriptsize{person,bicycle,bottle}}
&\textcolor{blue}{\scriptsize{chair,person,dining table}}
&\textcolor{blue}{\scriptsize{person,bottle,tvmonitor}}
&\textcolor{blue}{\scriptsize{cow,person,sofa}}
&\textcolor{blue}{\scriptsize{car,person,sofa}}
&\textcolor{blue}{\scriptsize{tvmonitor,chair,sofa}} \\ \hdashline
 & & & & &\\
\includegraphics[width=28mm,height=18mm]{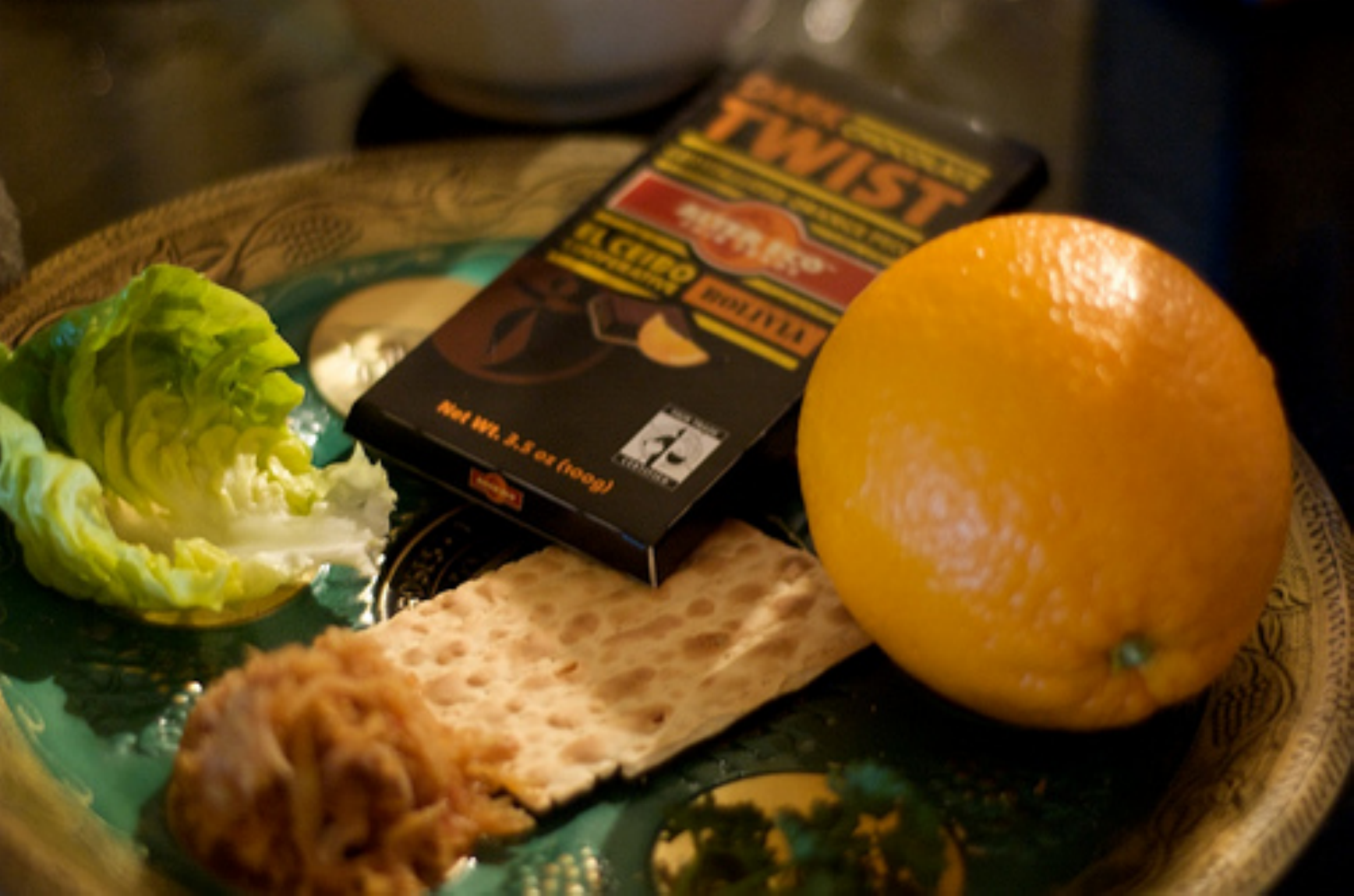}
&\includegraphics[width=28mm,height=18mm]{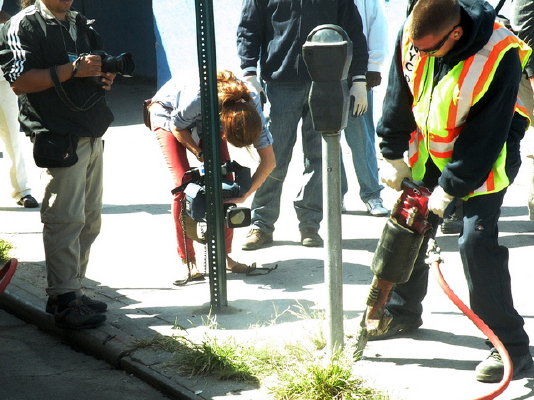}
&\includegraphics[width=28mm,height=18mm]{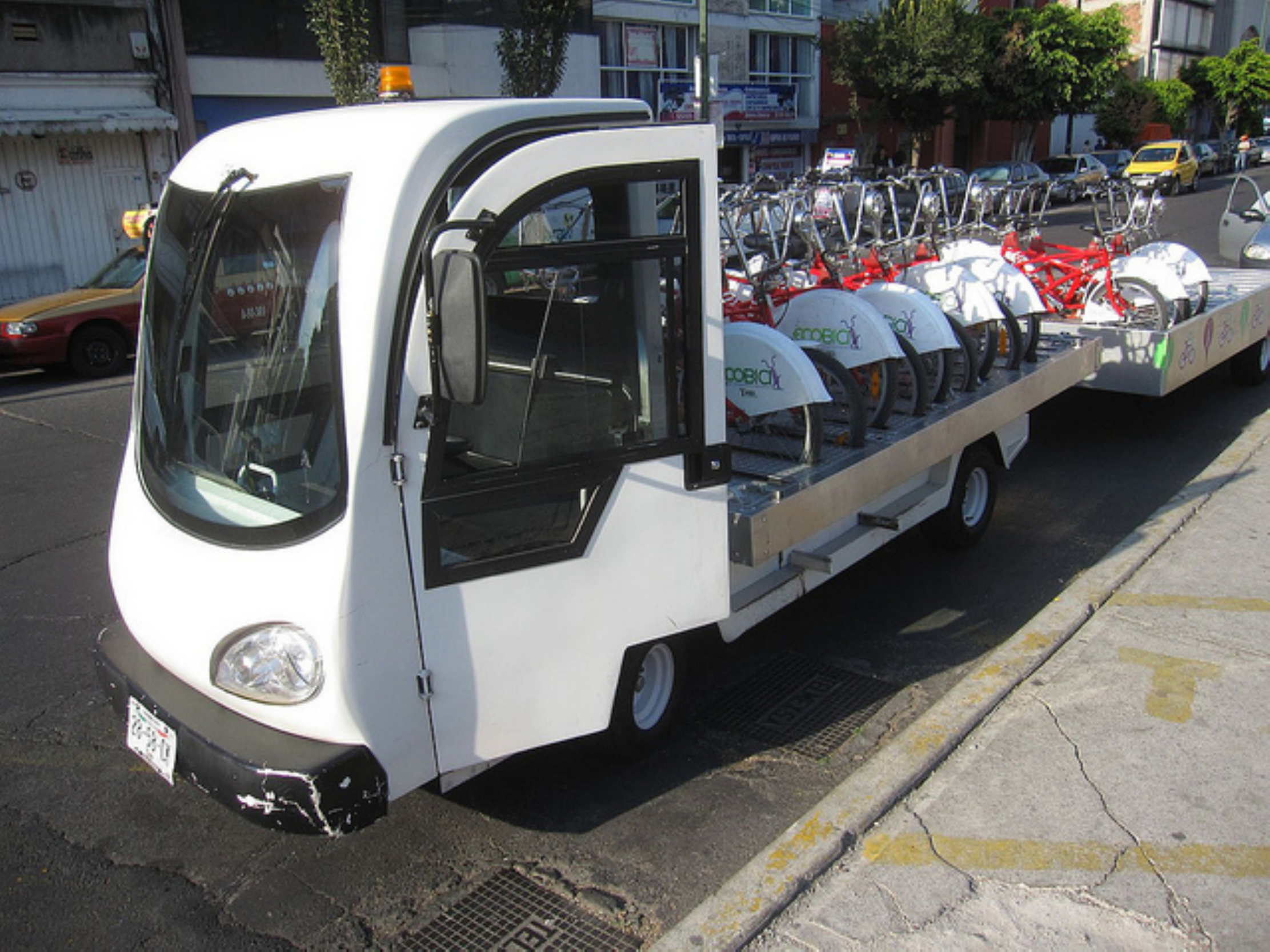}
&\includegraphics[width=28mm,height=18mm]{Figure20.pdf}
&\includegraphics[width=28mm,height=18mm]{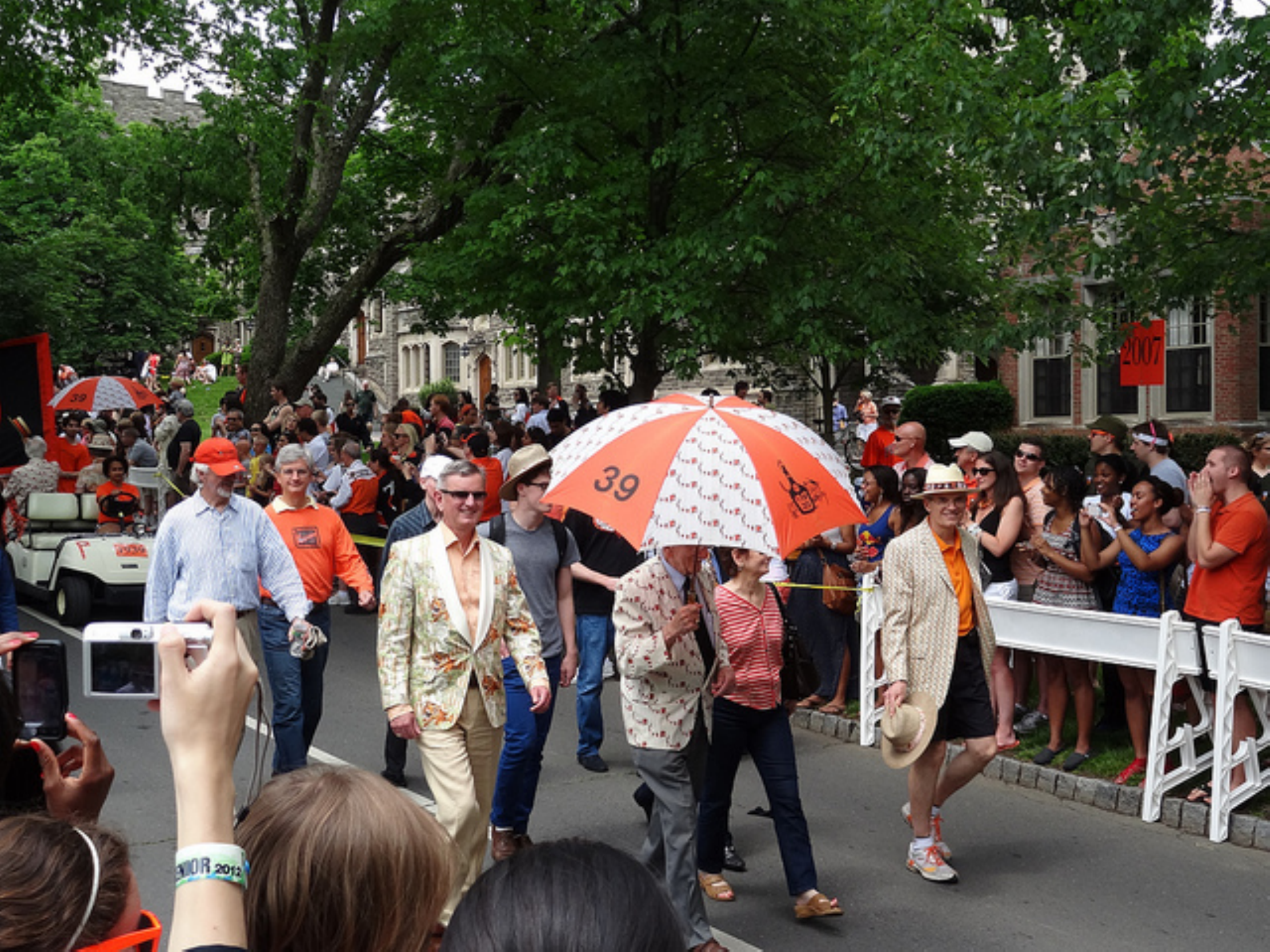}
&\includegraphics[width=28mm,height=18mm]{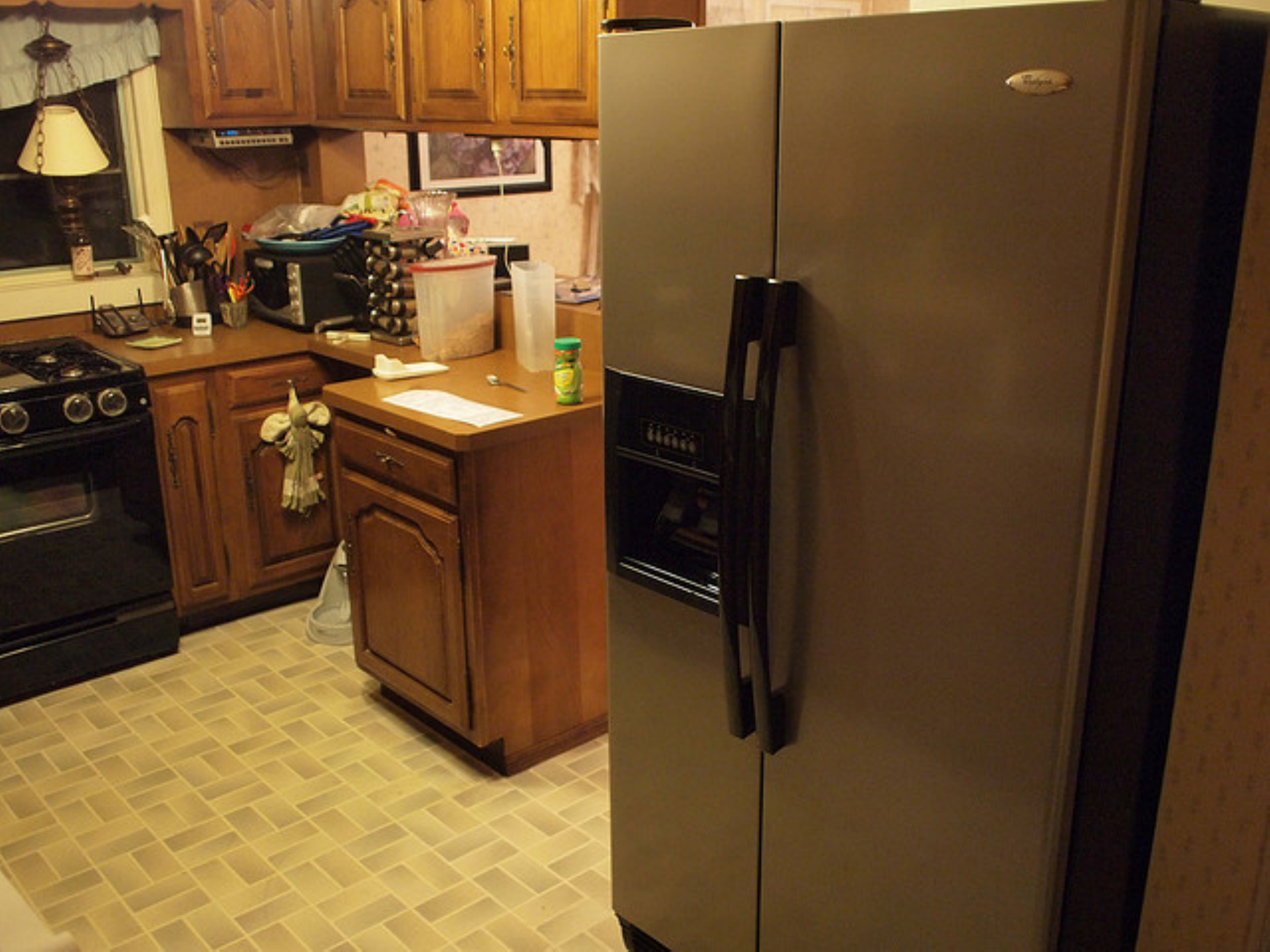}\\
\textcolor{red}{\scriptsize{orange,bowl,dining table}} 
&\textcolor{red}{\scriptsize{person,handbag,parking meter}}
&\textcolor{red}{\scriptsize{truck,bicycle,car}}
&\textcolor{red}{\scriptsize{pizza,dining table,spoon}} 
&\textcolor{red}{\scriptsize{person,umbrella,handbag}}
&\textcolor{red}{\scriptsize{spoon,microwave,refrigerator}}\\
\textcolor{blue}{\scriptsize{orange,bowl,book}} 
&\textcolor{blue}{\scriptsize{person,skis,backpack}} 
&\textcolor{blue}{\scriptsize{truck,car,person}} 
&\textcolor{blue}{\scriptsize{dining table,bowl,sandwich}} 
&\textcolor{blue}{\scriptsize{person,umbrella,backpack}} 
&\textcolor{blue}{\scriptsize{refrigerator,sink,oven}}\\ \hline
\end{tabular}}
\caption{Selected examples of the top-3 returned labels by ML-SGM (\textcolor{red}{red} word) and baseline ResNet101(\textcolor{blue}{blue} word) on the VOC 2007 dataset (first row) and MS-COCO dataset (last row). Best viewed in color and zoom in.}
\label{visual}
\vspace{-3mm}
\end{figure*}

\paragraph{\textbf{Comparison on MS-COCO}} As shown in Table \ref{coco}, ML-SGM achieves the best mAP, CF1, and OF1 on the MS-COCO dataset, which are all the most important performance indicators. In general, the most used input size is $448\times448$ in the training and inference phase. But it is worth noting that some methods evaluate their performance on different resolutions in the above two stages, \textit{e.g.}, C-Tran \cite{lanchantin2021general} and ADD-GCN \cite{ye2020attention}. For fair comparisons, we follow their resolution settings and present three results, which reach a new performance ranking with obvious advantages.  

Concretely, the CF1 and OF1 of ML-SGM in Top-3 case are 77.0\% and 79.3\%. These numbers in All cases are 80.4\% and 83.0\%. Compared with the plain RestNet101 model, ML-SGM gets 9.0\% improvements in mAP which validates the effectiveness of our proposed method. Moreover, ML-SGM conformably executes better than other comparing approaches, i.g. DSDL, SSGRL and ADD-GCN. The gains over the above methods in mAP are 3.4\%, 2.7\% and 0.8\% respectively. 
Next, we choose some images at random from different scenes and show top-3 predicted labels via ML-SGM and baseline ResNet-101 (Figure \ref{visual}) to visually understand the effectiveness of our model.

Furthermore, it is worth mentioning that F1 is the harmonic mean of precision (P) and recall (R). Ideally, we would expect both P and R to be high, but they naturally compete with each other. The ultimate goal is to achieve a better trade-off between P and R. However, the previous methods are less favored in such metrics. For example, SSGRL \cite{chen2019learning} outperforms our method on CP by 1.1\%, but we are able to be 7.3\% higher on CR. ADD-GCN \cite{ye2020attention} achieves 0.3\% higher on CR, but we outperform it by 3.7\% in terms of CP. It indicates that our approach may lose some performance on one metric but achieve significantly better on another. Overall, our approach preserves a better trade-off.

We analyze the performance of the proposed method on all the metrics in depth, especially on the precision (CP) and recall (CR). Then we find that our model can achieve higher CP when the threshold $\gamma$ is increased, since the class activation maps (instance regions) with higher prediction scores will be selected. In contrast, our model can achieve higher CR when the threshold $\gamma$ is decreased, since more class activation maps and instance regions will be selected. In summary, our proposed method is a multi-instance multi-label learning framework, and the selection of instance regions is crucial to the performance of multi-label image classification.

\paragraph{\textbf{Comparison on NUS-WIDE} } We further compare ML-SGM to the state-of-the-art methods such as SRN \cite{zhu2017learning}, CNN-RNN \cite{wang2016cnn}, PLA \cite{yazici2020orderless}, MLIC-KD-WSD \cite{liu2018multi}, ResNet-101 \cite{he2016deep}, CMA \cite{you2020cross} and \cite{chen2022sst} on the NUS-WIDE validation set. Table \ref{nus-wide} presents the quantitative results of all models. It can be clearly seen that ML-SGM not only can effectually learn from the large-scale dataset, but also outperforms the competition in most metrics, with 64.6\% mAP, 62.4\% CF1, 72.5\% (71.7\%) OF1 (Top-3) respectively.

\subsection{Results on Multi-Label Recognition with Partial Labels}
\paragraph{\textbf{Experimental Settings}} We follow previous works \cite{chen2022structured} to conduct experiments on Pascal VOC 2007 \cite{everingham2010pascal}, MS-COCO \cite{lin2014microsoft}, and Visual Genome 200 \cite{krishna2017visual} datasets for MLIR-PL evaluation. Since the three datasets are fully annotated, we randomly drop some labels to create the training set with partial labels. In this work, the proportion of dropped labels varies from 10\% to 90\% with the step of 10\%, resulting in 90\% to 10\% observed labels. For a fair comparison, we adopt the mean average precision (mAP) over all categories for evaluation under different proportions of known labels. And we also compute the average mAP over all proportions. 

\begin{table*}[htbp]
\caption{Performance of our ML-SGM and current state-of-the-art methods for MLIR-PL on VOC 2007, MS-COCO and Visual Genome 200 datasets. The \textcolor{red}{red} numbers indicate the best results, and the \textcolor{blue}{blue} numbers indicate the sub-optimal results. * denotes the performance of our implementation.}
\begin{center}                                                     
\setlength{\tabcolsep}{2.2mm}{
\begin{tabular}{c|c|c|c|c|c|c|c|c|c|c|c}
\toprule[1pt]
\textbf{Datasets} &\textbf{Methods} &\textbf{10\%} &\textbf{20\%} &\textbf{30\%} &\textbf{40\%} &\textbf{50\%} &\textbf{60\%} &\textbf{70\%} &\textbf{80\%} &\textbf{90\%} &\textbf{Ave. mAP}  \\ \hline\hline 
\multirow{5}{*}{Pascal VOC 2007} &SSGRL* \cite{chen2019learning} &77.7 &87.6 &89.9 &90.7 &91.4 &91.8 &92.0 &92.2 &92.2 &89.5\\
                                 &ML-GCN* \cite{chen2019multi} &74.5 &87.4 &89.7 &90.8 &91.0 &91.3 &91.5 &91.8 &92.0 &88.9\\
                                 &KGGR* \cite{chen2020knowledge} &81.3 &88.1 &89.9 &90.4 &91.2 &91.4 &91.5 &91.6 &91.8 &89.7\\
                                 &Curriculum labeling* \cite{durand2019learning} &44.7 &76.8 &88.6 &90.2 &90.7 &91.1 &91.6 &91.7 &91.9 &84.1\\
                                 &Partial BCE* \cite{durand2019learning} &80.7 &88.4 &89.9 &90.7 &91.2 &91.8 &92.3 &92.4 &92.5 &90.0\\
                                 &SST \cite{chen2022structured} &\textcolor{blue}{81.5} &\textcolor{blue}{89.0} &\textcolor{blue}{90.3} &\textcolor{blue}{91.0} &\textcolor{blue}{91.6} &\textcolor{blue}{92.0} &\textcolor{blue}{92.5} &\textcolor{blue}{92.6} &\textcolor{blue}{92.7} &90.4\\
                                  &SARB \cite{pu2022semantic} &- &- &- &- &- &- &- &- &- &\textcolor{blue}{90.7}\\
                                 &Ours &\textcolor{red}{86.3} &\textcolor{red}{89.5} &\textcolor{red}{91.3} &\textcolor{red}{91.8} &\textcolor{red}{92.6} &\textcolor{red}{92.9} &\textcolor{red}{93.0} &\textcolor{red}{93.3} &\textcolor{red}{93.2} &\textcolor{red}{91.6}\\\hline\hline 
\multirow{5}{*}{MS-COCO}         &SSGRL* \cite{chen2019learning} &62.5 &70.5 &73.2 &74.5 &76.3 &76.5 &77.1 &77.9 &78.4 &74.1\\
                                 &ML-GCN* \cite{chen2019multi} &63.8 &70.9 &72.8 &74.0 &76.7 &77.1 &77.3 &78.3 &78.6 &74.1\\
                                 &KGGR* \cite{chen2020knowledge} &66.6 &71.4 &73.8 &76.7 &77.5 &77.9 &78.4 &78.7 &79.1 &75.6\\
                                 &Curriculum labeling* \cite{durand2019learning} &26.7 &31.8 &51.5 &65.4 &70.0 &71.9 &74.0 &77.4 &78.0 &60.7\\
                                 &Partial BCE* \cite{durand2019learning} &61.6 &70.5 &\textcolor{blue}{74.1} &76.3 &77.2 &77.7 &78.2 &78.4 &78.5 &74.7\\
                                 &SST \cite{chen2022structured} &\textcolor{blue}{68.1} &\textcolor{blue}{73.5} &\textcolor{blue}{75.9} &\textcolor{blue}{77.3} &\textcolor{blue}{78.1} &\textcolor{blue}{78.9} &\textcolor{blue}{79.2} &\textcolor{blue}{79.6} &\textcolor{blue}{79.9} &76.7\\
                                  &SARB \cite{pu2022semantic} &- &- &- &- &- &- &- &- &- &\textcolor{blue}{77.9}\\
                                 &Ours &\textcolor{red}{72.3} &\textcolor{red}{75.6} &\textcolor{red}{77.8} &\textcolor{red}{79.5} &\textcolor{red}{80.2} &\textcolor{red}{80.8} &\textcolor{red}{81.4} &\textcolor{red}{81.7} &\textcolor{red}{82.0} &\textcolor{red}{79.0}\\\hline\hline  
\multirow{5}{*}{VG-200}        &SSGRL* \cite{chen2019learning} &34.6 &37.3 &39.2 &40.1 &40.4 &41.0 &41.3 &41.6 &42.1 &39.7\\
                                 &ML-GCN* \cite{chen2019multi} &32.0 &37.8 &38.8 &39.1 &39.6 &40.0 &41.9 &42.3 &42.5 &39.3\\
                                 &KGGR* \cite{chen2020knowledge} &36.0 &40.0 &41.2 &41.5 &42.0 &42.5 &43.3 &43.6 &43.8 &41.5\\
                                 &Curriculum labeling* \cite{durand2019learning} &12.1 &19.1 &25.1 &26.7 &30.0 &31.7 &35.3 &36.8 &38.5 &28.4\\
                                 &Partial BCE* \cite{durand2019learning} &27.4 &38.1 &40.2 &40.9 &41.5 &42.1 &42.4 &42.7 &42.7 &39.8\\
                                 &SST \cite{chen2022structured} &\textcolor{blue}{38.8} &\textcolor{blue}{39.4} &\textcolor{blue}{41.1} &\textcolor{blue}{41.8} &\textcolor{blue}{42.7} &\textcolor{blue}{42.9} &\textcolor{blue}{43.0} &\textcolor{blue}{43.2} &\textcolor{blue}{43.5} &41.8\\
                                 &SARB \cite{pu2022semantic} &- &- &- &- &- &- &- &- &- &\textcolor{blue}{45.6}\\
                                 &Ours &\textcolor{red}{43.6} &\textcolor{red}{43.9} &\textcolor{red}{45.2} &\textcolor{red}{45.8} &\textcolor{red}{46.6} &\textcolor{red}{47.0} &\textcolor{red}{47.3} &\textcolor{red}{47.5} &\textcolor{red}{47.8} &\textcolor{red}{46.1} \\
\bottomrule[1pt]
\end{tabular}}
\label{partial-label}
\end{center}
\vspace{-5mm}
\end{table*}

\paragraph{\textbf{Comparison on Pascal VOC 2007}} We report the performance comparisons on VOC 2007 dataset in Table \ref{partial-label}. Since this dataset merely covers 20 categories and it is more simple than MS-COCO and Visual Genome, current methods achieve comparable results when keeping a certain proportion of known labels (\textit{e.g.}, more than 40\%). But their performances drop dramatically when the proportion decreases to 10\% and 20\%. Our ML-SGM also suffers a performance drop, but it consistently outperforms current methods for all proportion settings. Specifically, it outperforms SST and KGGR by 4.8\% and 5.0\% when known labels are merely 10\%.

\paragraph{\textbf{Comparison on MS-COCO}} Next, we present the comparison results on the MS-COCO dataset in Table \ref{partial-label}. We find the traditional multi-label recognition methods SSGRL and GCN-ML can achieve competitive performance when the proportion of known labels is high (\textit{e.g.}, 70\%-90\%), but suffer from an obvious performance drop when the proportion decreases. By introducing category correlations and instance-label matching relationships to transfer knowledge of known labels into unknown labels, our model achieves the best performance over all known label proportion settings. It is noteworthy that our ML-SGM obtains more significant performance improvement compared to existing methods when decreasing the known label proportions. For example, the mAP improvements over the previous second-best SST model are 2.1\% and 4.2\% when using 90\% and 10\% known labels, respectively. These results indicate that the proposed method is very effective in handling missing labels.

\paragraph{\textbf{Comparison on Visual Genome 200}} As previously discussed, VG-200 is a more challenging benchmark that covers much more categories. We report the performance comparisons in Table \ref{partial-label}. As shown in Table \ref{partial-label}, existing methods achieve quite poor performances, but our ML-SGM framework obtains the best performance over all proportion settings. Specifically, its average mAP is 46.1\%, outperforming the best SARB algorithm by 0.5\%. In addition, it outperforms leading multi-label image classification methods SST and KGGR by 4.8\% and by 7.6\% when known labels are 10\%.

\subsection{Results on Multi-Label Few-Shot Learning}
\paragraph{\textbf{Experimental Settings}} We follow previous works \cite{alfassy2019laso,chen2020knowledge} to conduct experiments on MS-COCO \cite{lin2014microsoft} and Visual Genome 500\cite{krishna2017visual} datasets for ML-FSL evaluation. For MS-COCO benchmark, we split the 80 categories into 64 base categories and 16 novel categories. Specifically, the novel categories are \textit{bicycle}, \textit{boat}, \textit{stop sign}, \textit{bird}, \textit{backpack}, \textit{frisbee}, \textit{snowboard}, \textit{surfboard}, \textit{cup}, \textit{fork}, \textit{spoon}, \textit{broccoli}, \textit{chair}, \textit{keyboard}, \textit{microwave}, and \textit{vase}. For Visual Genome 500 benchmark, we randomly split VG-500 into 400 base categories and 100 novel categories. To ensure fair comparisons with current works, we utilize the trainval samples of the base categories as the base set and randomly select K (K=1,5) trainval samples from the novel categories as the novel set. We evaluate the models on the test samples of the novel categories. 

\paragraph{\textbf{Comparison on MS-COCO}} We first compare our proposed method with the state-of-the-art method SSGRL \cite{chen2020knowledge} and LaSO \cite{alfassy2019laso} on MS-COCO dataset. The results are reported in Table \ref{few-shot}. As shown, LaSO and SSGRL obtain mAPs of 48.4\% (60.8) and 52.3\% (63.5) on the 1-shot (5-shot) setting, respectively. In contrast, our ML-SGM incorporates instance spatial relationships and label global correlations to guide feature and semantic propagation among the different categories, leading to superior performance. Especially, our model obtains mAPs of 54.1\% and 65.8\% on the 1-shot and 5-shot settings, outperforming SSGRL by 1.8\% and 2.3\%. 

\paragraph{\textbf{Comparison on Visual Genome 500}} Visual Genome 500 is a more complex and realistic benchmark for the ML-FSL problem since it covers a wider range of categories.
Thus we utilize it as a new evaluation benchmark and present the comparison results in \ref{few-shot}. As shown in Figure \ref{few-shot}, our proposed framework clearly outperforms the LaSO and SSGRL by a sizable margin. Specifically, our ML-SGM achieves mAPs of 23.6\% and 31.9\%, yielding about 7.0\% and 10.1\% improvement over LaSO on the 1-shot and 5-shot settings, respectively. Besides, it outperforms the state-of-the-art ML-FSL method SSGRL by 2.9\% and 5.8\% on the 1-shot and 5-shot settings.   

\begin{table}[t]
\caption{Performance of our ML-SGM and current state-of-the-art competitors for MLIR-FSL on MS-COCO and Visual Genome 500 datasets. * denotes the performance of our implementation.}
\begin{center}
\setlength{\tabcolsep}{3.2mm}{
\begin{tabular}{c|c|c|c}
\toprule[1pt]
\textbf{Datasets} &\textbf{Methods} &\textbf{1-shot} &\textbf{5-shot} \\ \hline\hline 
\multirow{3}{*}{MS-COCO}        &LaSO(ResNet-101)* \cite{alfassy2019laso} &48.4 &60.8 \\
                                 &KGGR \cite{chen2020knowledge} &\textcolor{blue}{52.3} &\textcolor{blue}{63.5} \\
                                 &Ours &\textcolor{red}{54.1} &\textcolor{red}{65.8} \\\hline\hline  
\multirow{3}{*}{VG-500}        &LaSO(ResNet-101)* \cite{alfassy2019laso} &16.6 &21.8\\                                 &KGGR \cite{chen2020knowledge} &\textcolor{blue}{20.7} &\textcolor{blue}{26.1}\\
                                 &Ours &\textcolor{red}{23.6} &\textcolor{red}{31.9}\\
\bottomrule[1pt]
\end{tabular}}
\label{few-shot}
\end{center}
\vspace{-5mm}
\end{table}

\subsection{Performance Analysis}
\paragraph{\textbf{Ablation Studies}} To evaluate the effectiveness of each component in our proposed framework, we reconstruct our model with different ablation factors on the MS-COCO and VOC 2007 datasets. The performance of different variant models are shown in Table \ref{ablation_studies}. Since our proposed framework builds on ResNet-101 \cite{he2016deep}, we first compare it with this baseline to analyze the contributions of our ML-SGM framework (Index-1). Specifically, we simply replace the last fully connected layer of ResNet-101 with a 2,048-to-$C$ fully connected layer and then use sigmoid functions to predict the probability of each category. And the mAP respectively drops 7.8\% and 5.3\% on the MS-COCO and VOC2007 datasets, which demonstrates the necessity of the proposed modules to obtain the best classification results. It can be found that only adopting content-aware semantic activation module for instance generation can significantly improve the performance (Index-2), \textit{e.g.}, 0.6\% on MS-COCO, which illustrates the effectiveness of exploiting content-aware category features for MLIR task. Besides, as objects normally co-occur in an image, it is desirable to model category dependencies to improve the classification performance. Our proposed instance spatial graph $\mathbb{G}^o$ could enhance image-special category correlations and suppress background noises, which further boost performance by 1.2\% on MS-COCO (Index-3).

\begin{figure*}[t]
\centering  
\centerline{\includegraphics[scale=0.45]{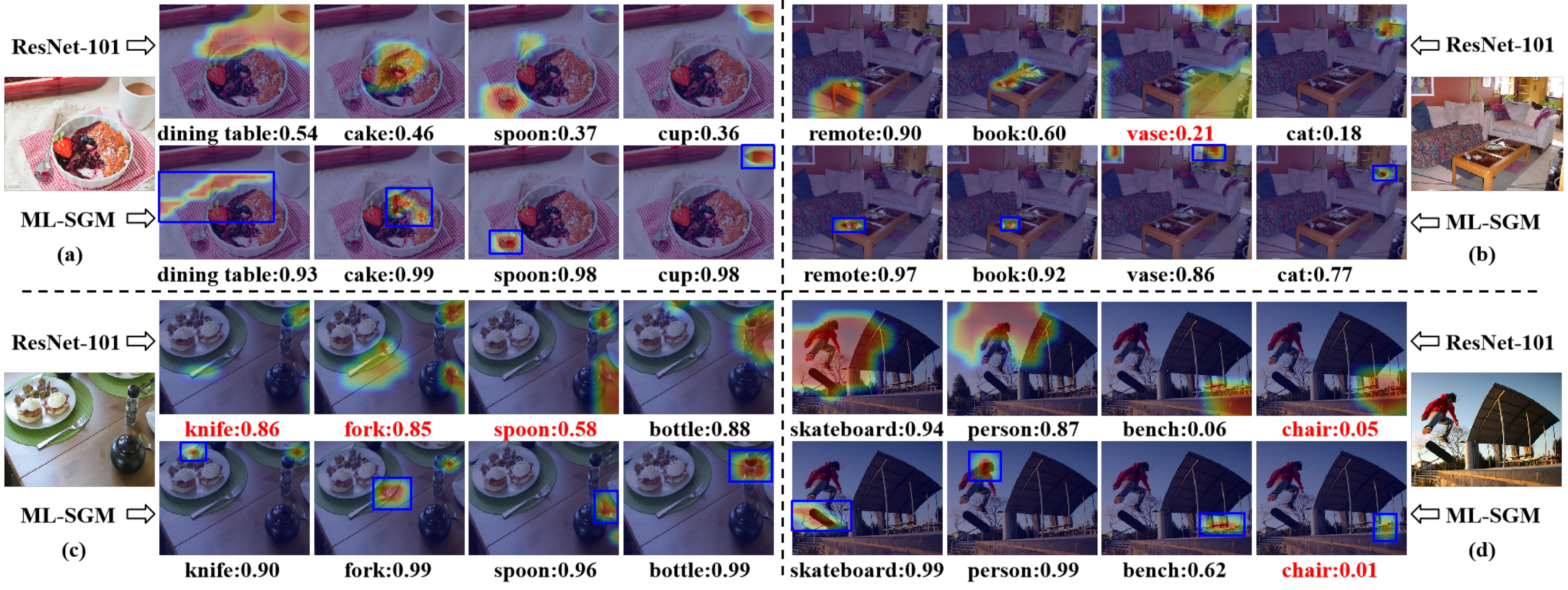}}
\caption{Visualization analyses of ResNet-101 and our proposed ML-SGM. We present several labels and their corresponding location boxes and the incorrect label localization in the image are highlighted in red. Compared to the baseline ResNet-101 our ML-SGM has the ability to handle objects in small scales in a) and b) or with confused appearances in c) and d).}       
\label{activation_map}
\end{figure*}

\begin{table}
\caption{Ablation study for different components of our framework. $\mathcal{M}_{BL}$ denotes the baseline ResNet101. $\mathcal{M}_{CASC}$ denotes content-aware semantic activation module. $\mathcal{M}_{\mathbb{G}^o}$ denotes instance spatial graph. $\mathcal{M}_{\mathbb{G}^l}$ denotes label semantic graph. $\mathcal{M}_{ILME}$ denotes instance-label matching edges.}     
\begin{center}
\setlength{\tabcolsep}{0.20mm}{
\begin{tabular}{c|ccccc|cc}
\hline
\multirow{2}{*}{\textbf{Index}} &\multirow{2}{*}{$~\bm{\mathcal{M}}_{\bm{BL}}~$} &\multirow{2}{*}{$\bm{\mathcal{M}}_{\bm{CASC}}$} &\multirow{2}{*}{ $~~\bm{\mathcal{M}}_{\bm{\mathbb{G}^o}}~~$} &\multirow{2}{*}{$~~\bm{\mathcal{M}}_{\bm{\mathbb{G}^l}}~~$} &\multirow{2}{*}{$\bm{\mathcal{M}_{ILME}}$} &\multicolumn{2}{c}{\textbf{mAP}} \\
 & & & & & &\textbf{COCO} & \textbf{VOC}\\ \hline \hline
\rule{0pt}{8pt} 1 &\ding{51} &\ding{55} &\ding{55} &\ding{55} &\ding{55} &77.3 &89.9 \\  
\rule{0pt}{8pt} 2 &\ding{51} &\ding{51} &\ding{55} &\ding{55} &\ding{55} &78.0 &90.5 \\ 
\rule{0pt}{8pt} 3 &\ding{51} &\ding{51} & \ding{51} &\ding{55} &\ding{55} &79.2 &92.0 \\ 
\rule{0pt}{8pt} 4 &\ding{51} &\ding{55} &\ding{55} &\ding{51} &\ding{55} &78.4 &91.2 \\ 
\rule{0pt}{8pt} 5 &\ding{55} &\ding{55} &\ding{51} &\ding{51} &\ding{55} &81.1 &93.2 \\ 
\rule{0pt}{8pt} 6 &\ding{51} &\ding{51} &\ding{51}&\ding{51} &\ding{55} &82.6 &94.3 \\
\rule{0pt}{8pt} 7 &\ding{51} &\ding{51} &\ding{51} &\ding{51} &\ding{51} &85.1 &95.2 \\ \hline
\end{tabular}}
\label{ablation_studies}
\end{center}
\vspace{-5mm}
\end{table} 

\begin{table}[t]
\caption{The computation resource of each module in our proposed model, including the number of parameters (params) and floating point operations (FLOPs).}
\begin{center}
\setlength{\tabcolsep}{8.5mm}{
\begin{tabular}{c|cc}
\toprule[1pt]
\textbf{Methods} 
&\textbf{\#params(M)} &\textbf{FLOPs} \\ \hline\hline 
ResNet101   &42.51   &31.53 \\
+CSAC      &42.52   &31.57 \\
+Encoder    &44.89   &32.26 \\
+GCL        &57.62   &36.09  \\ 
+Decoder    &57.65   &36.11 \\ 
\bottomrule[1pt]
\end{tabular}}
\label{complexity}
\end{center}
\vspace{-5mm}
\end{table}

We also construct the label semantic graph $\mathbb{G}^l$ to capture label global correlations which are inferred from knowledge beyond a single image. As shown in the Index-4 of Table \ref{ablation_studies}, Using label semantic relationships alone can also bring performance improvements, \textit{e.g.}, 1.1\% on MS-COCO. And the performance further boosts for 2.7\% by jointing instance spatial graph $\mathbb{G}^o$ (Index-5), which illustrates that simultaneously modeling the spatial dependencies and semantic co-occurrences among categories are essential for multi-label image understanding. Finally, to verify the effectiveness of instance-label matching edges, we replace these matching edges with a simple concatenation operation. Specifically, we train instance spatial graph $\mathbb{G}^o$ and label semantic graph $\mathbb{G}^l$ in parallel using graph neural network, and then directly concatenate generated instances and label features to identify multiple labels. The performance shows a clear drop (2.5\%) in mAP on MS-COCO (Index-6 \& 7), which illustrates that the ILMS module plays a vital role in our model, and it could guide the model to learn representative category features via establishing explicit correspondence, and further improve the classification precision.     

In addition to the classification performance, we further explore the computational complexity of our method, including the number of parameters and FLOPs of each module in the proposed model. As reported in Table \ref{complexity}, the graph convolution layer in the ILMS module dominates the computation cost (\textit{i.e.}, 44.89M \textit{v.s.} 57.62M) of our ML-SGM. Reducing the number of graph convolution blocks would significantly decrease the complexity of the model. Moreover, our Encoder module requires approximately 2.37M learnable parameters to map the attributes of all nodes and edges into a latent feature space. It is worth noting that our CSAC and Decoder modules require very few parameters and enjoy the time complexity (measured in FLOPs) that is negligible compared to the feature extraction network ResNet101.

\begin{figure}[t]
\centering 
	\subfloat[The threshold $\gamma$ of instance.]{
		\includegraphics[width=40mm,height=25mm]{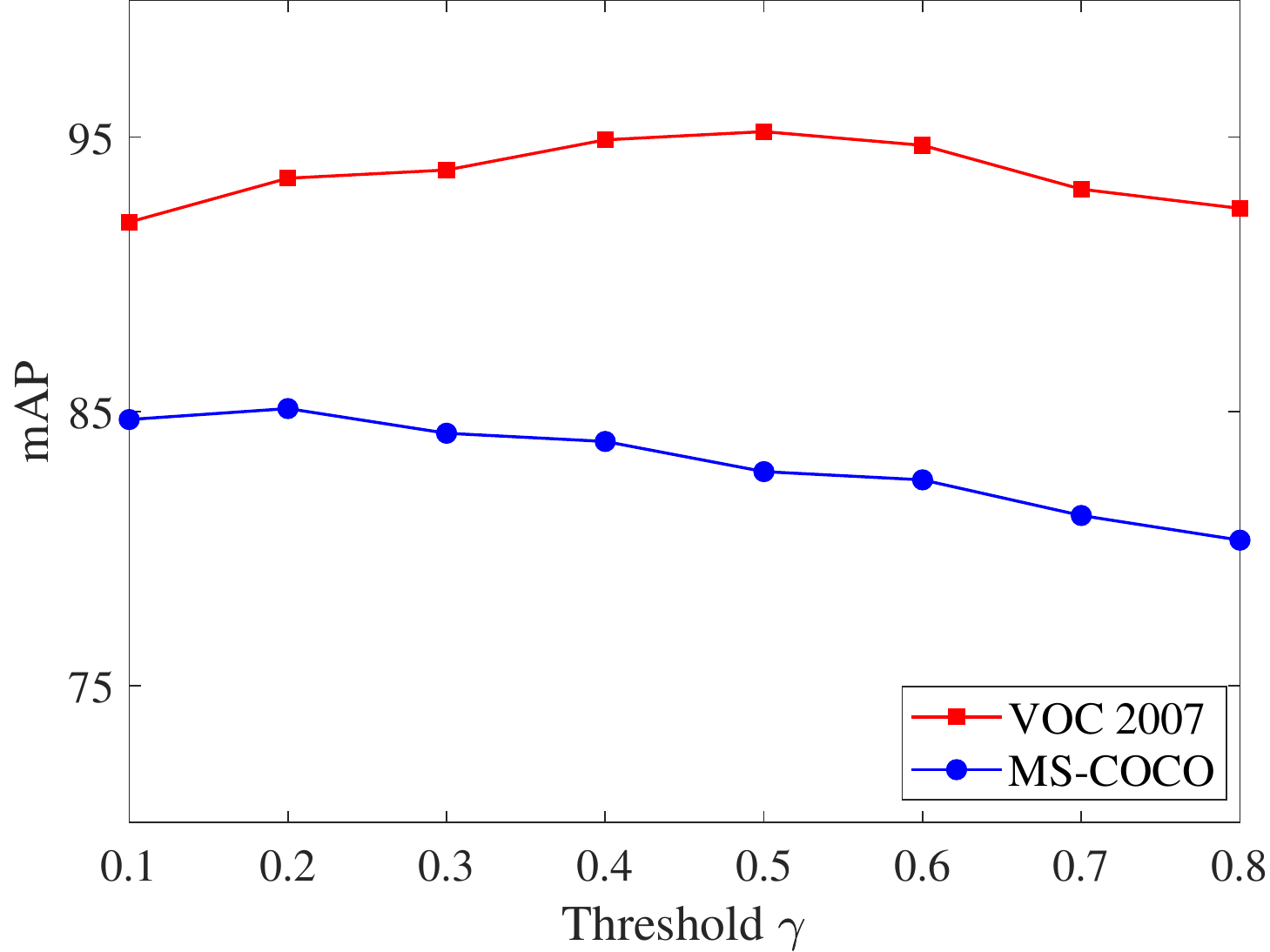}}
	\subfloat[The num $k$ of conv module.]{
		\includegraphics[width=40mm,height=25mm]{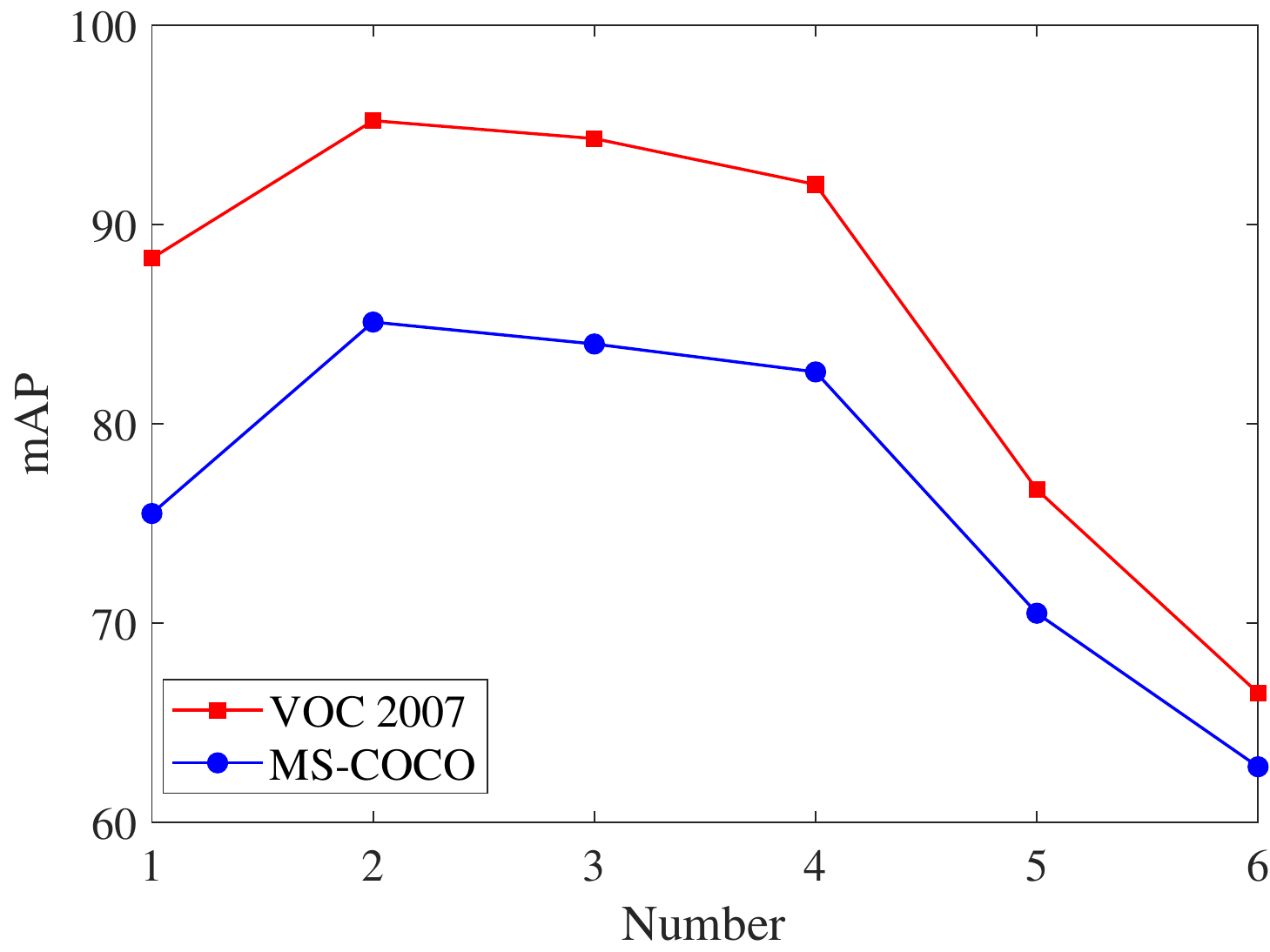}} \\
	\subfloat[The dim $d$ of conv module.]{
		\includegraphics[width=40mm,height=25mm]{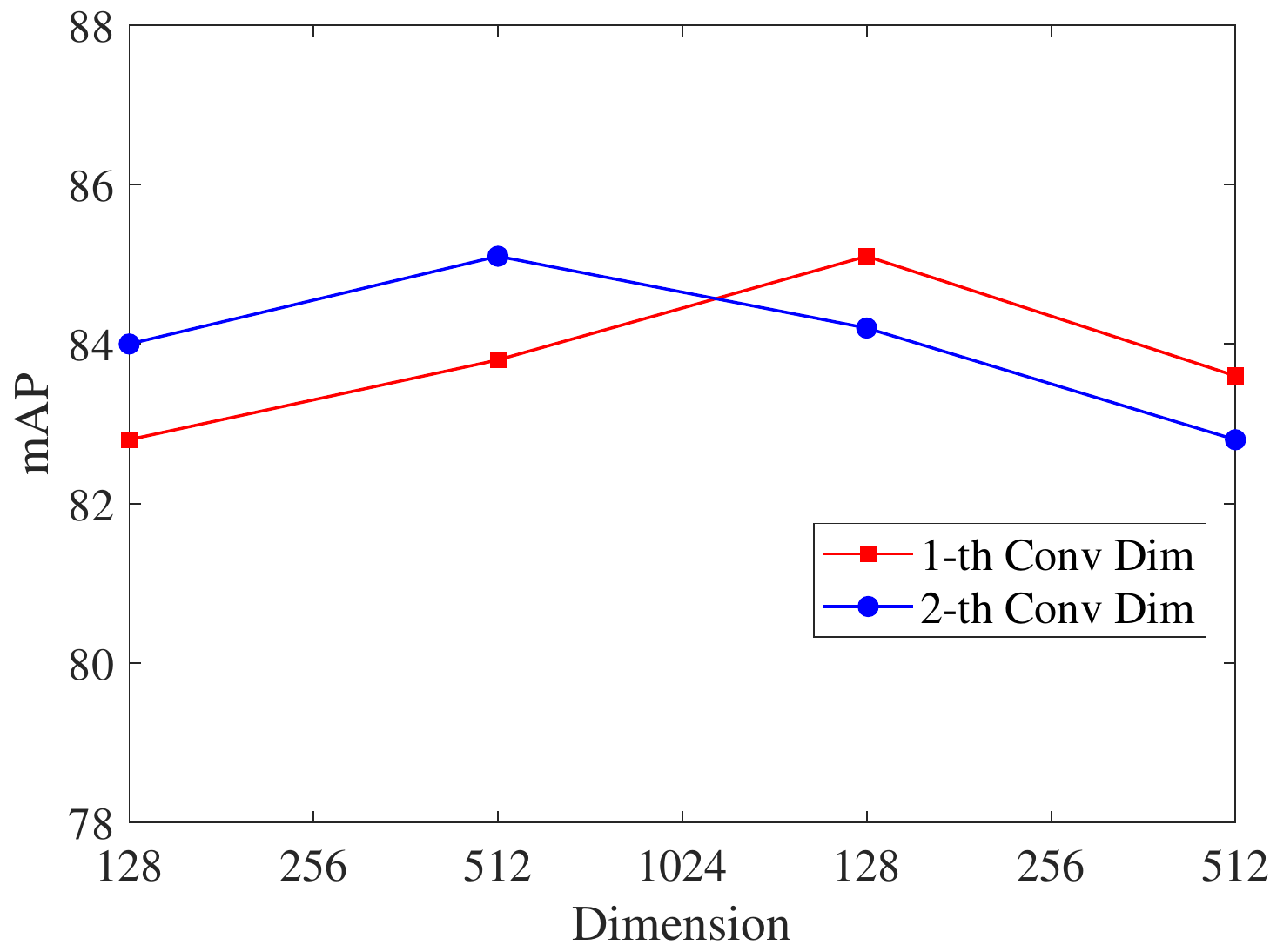}}
	\subfloat[The imbalance factor $\beta$.]{
		\includegraphics[width=40mm,height=25mm]{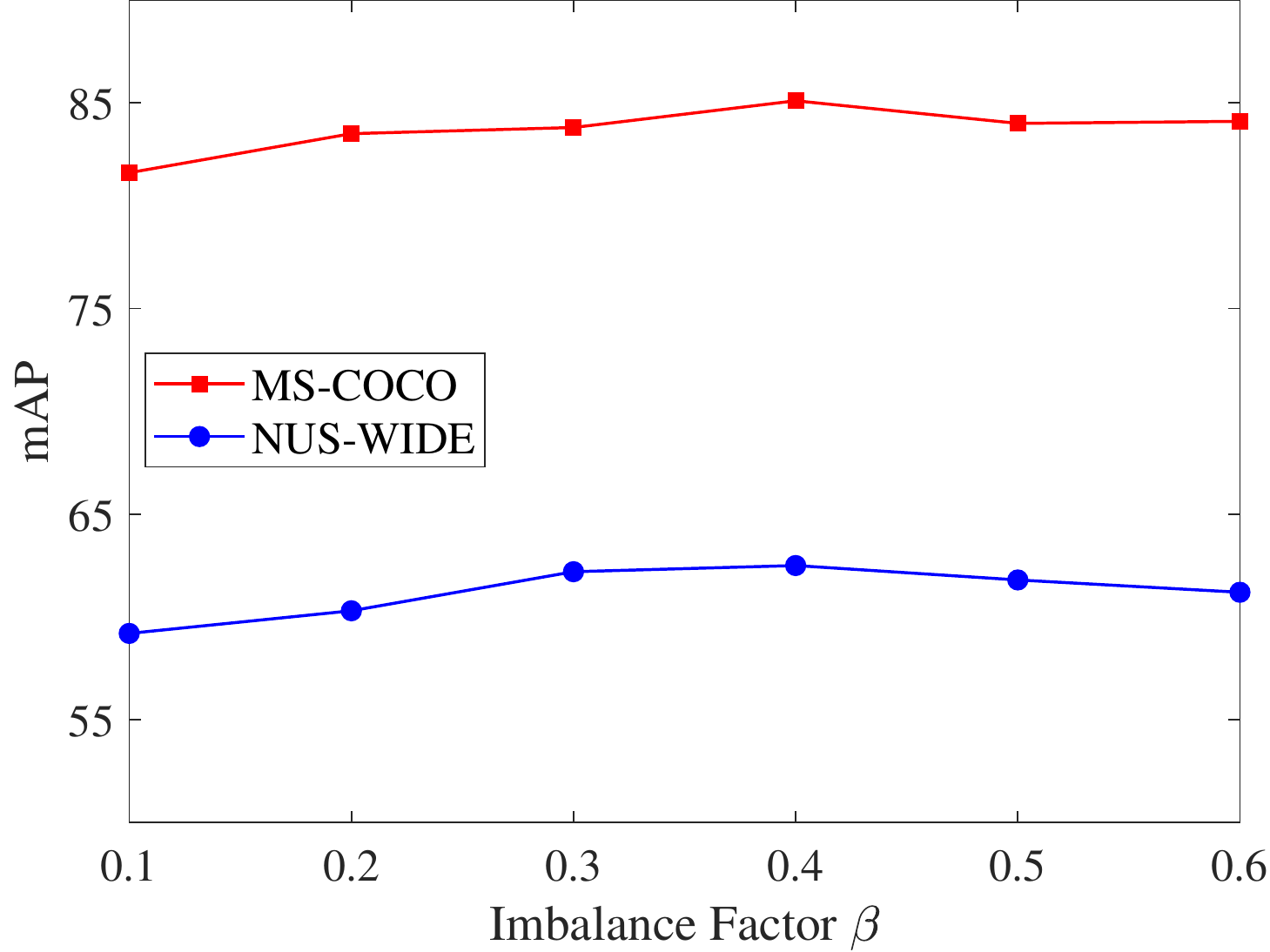}}
	\caption{Accuracy comparisons with different parameter values.}
	\label{parameter}
\vspace{-5mm}
\end{figure} 

\paragraph{\textbf{Hyper-parameter Sensitivity}} Furthermore, we study the sensitivity of our proposed method given different parameter settings. We first vary the threshold value $\gamma$ of the selected instances per image and show the results in Figure \ref{parameter} (a). Note that, if we keep all category instances, the model will no doubt contain a lot of noise instances, which is difficult to converge and increases computational complexity. However, when too many instances are filtered out, the accuracy drops since some positive instances in images may be removed incorrectly. Therefore, an appropriate threshold $\gamma$ is crucial for the accurate parsing of multi-label recognition. Empirically, the optimal value of $\gamma$ is set to 0.5 in VOC 2007 dataset, and 0.2 in both MS-COCO and NUS-WIDE datasets. Then, we show the performance results with different numbers of convolution modules $k$ for our model in Figure \ref{parameter} (b). As the number of convolution modules increases, the accuracy drops on all the evaluation datasets. Thus, we set $k=2$ in our experiments. Next, we study the sensitivity of ML-SGM with respect to its two parameters $d$ and $\beta$, which are the dimension of conv module and class imbalance factor, respectively. From Figure \ref{parameter} (c)-(d), we find that the $d$ value has a great influence on the performance of our ML-SGM. Meanwhile, our model is fairly robust to the class imbalance factor $\beta$. Additionally, we find that the performance of the proposed model is insensitive to changes in the $k$ value of $k$-Nearest Neighbor criterion. The main reason is that the multi-label datasets used in MLIR tasks usually contain few instances per image.

\begin{figure}[t]
\centering
\subfloat{
\begin{minipage}{0.5\textwidth}
\includegraphics[width=88mm,height=48mm]{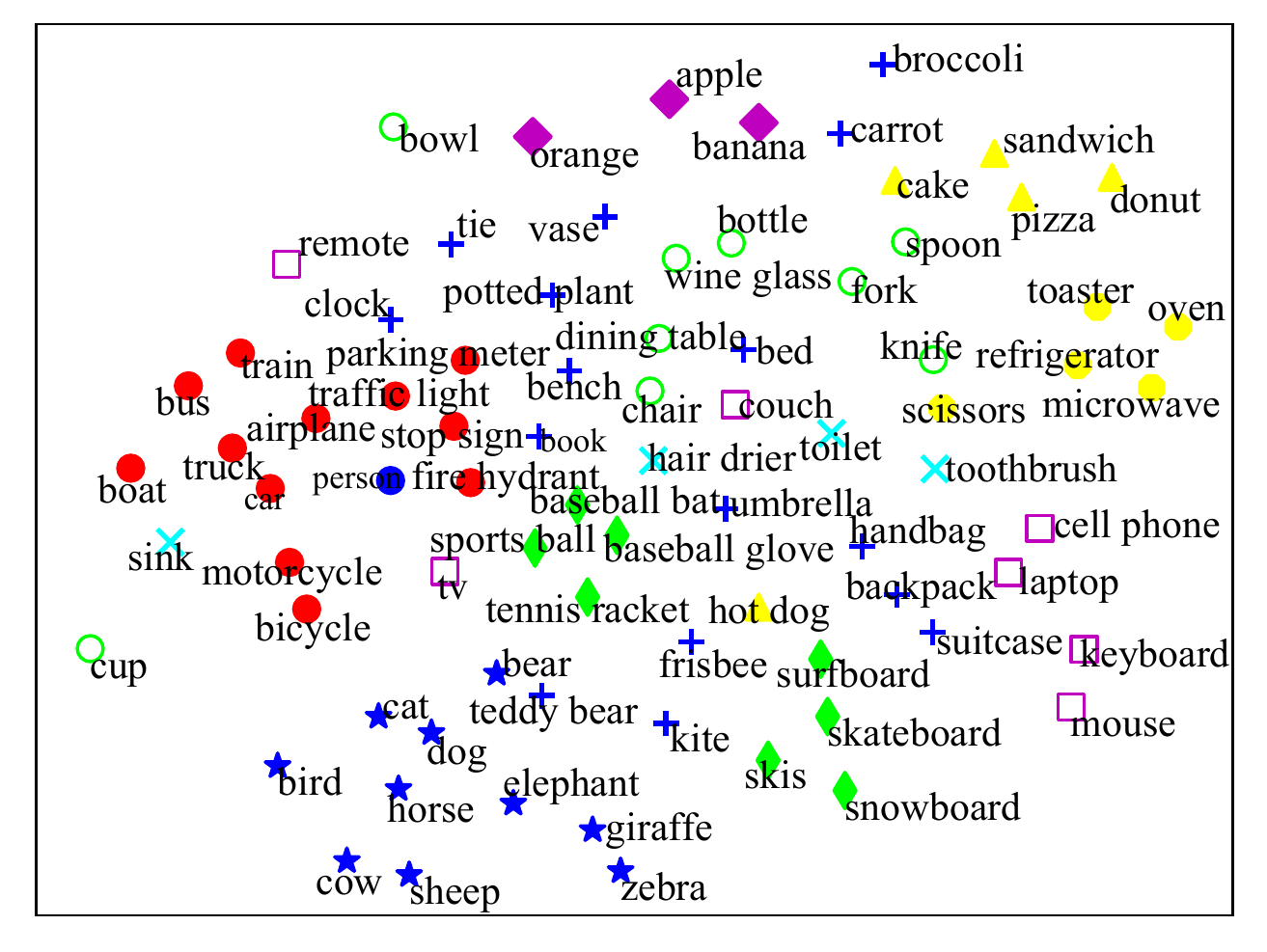}
\vspace{-1mm}
\end{minipage}}

\subfloat{
\begin{minipage}{0.5\textwidth}
\hspace{0.8mm}
\includegraphics[width=85mm,height=8mm]{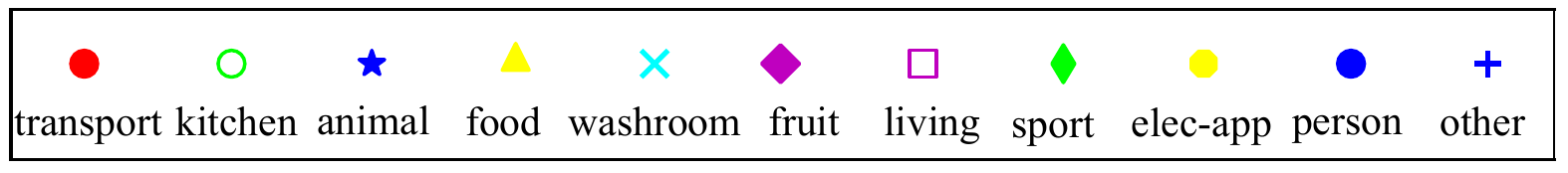}
\end{minipage}}
\caption{Visualization of the learned classifiers by our proposed model on the MS-COCO dataset.}
\label{label_cor}
\end{figure}

\paragraph{\textbf{Interpretable Visualization}} We adopt CAM \cite{zhou2016learning} to exhibit the visualization results of baseline ResNet-101 and the proposed ML-SGM in Figure \ref{activation_map}. Benefiting from the instance spatial graph $\mathbb{G}^o$ and label semantic graph $\mathbb{G}^l$, our model could capture more accurate localization and effectively perceive small object, \textit{e.g.}, \textit{spoon} in Figure \ref{activation_map} (a) as well as \textit{vase} , \textit{remote}, and \textit{book} in Figure \ref{activation_map} (b). Furthermore, since ResNet-101 could not distinguish objects with similar appearances, \textit{e.g.}, the triplet labels \{\textit{knife}, \textit{fork}, \textit{spoon}\} in Figure \ref{activation_map} (c) and the paired labels \{\textit{skateboard}, \textit{snowboard}\} and \{\textit{bench}, \textit{chair}\} in Figure \ref{activation_map} (d), these issues can be well handled by our proposed instance-label matching module benefiting from the instance-label explicit correspondence captured by graph neural network. In Figure \ref{label_cor}, we utilize \textit{t}-SNE \cite{van2008visualizing} to visualize the final label representations learned by our proposed method, which demonstrates that the meaningful semantic topology is maintained. Specifically, the learned representations exhibit clear cluster patterns. For example, the \textit{apple}, \textit{orange} and \textit{banana} within one super concept (\textit{e.g.} \textit{animal}) as well as \textit{person} and \textit{car} with high-frequent co-occurrences, tend to be close in the label semantic space. This is consistent with common sense, which further indicates that our learned label representations may not be limited to the corresponding dataset, but may enjoy generalization capacities. 

\begin{table}[t]
\caption{The comparison of inference speed and mAP on the MS-COCO dataset. All experiments are evaluated at a single 2080ti GPU.}
\begin{center}
\setlength{\tabcolsep}{8.5mm}{
\begin{tabular}{c|cc}
\toprule[1pt]
\textbf{Methods} 
&\textbf{FPS(Hz)} &\textbf{mAP(\%)} \\ \hline\hline 
ResNet101   &68.2  &77.3 \\
ML-GCN      &62.4  &83.0 \\
SSGRL       &34.3  &83.6  \\ 
ADD-GCN     &57.9  &83.9 \\ \hline\hline 
Our w/o CASC &54.3 &83.7   \\
Our w/o ILMS & 63.5  &82.6 \\
Our ML-SGM  &52.4  &85.1 \\
\bottomrule[1pt]
\end{tabular}}
\label{time}
\end{center}
\vspace{-5mm}
\end{table}

\paragraph{\textbf{Time Consumption}} To compare the computational time between our ML-SGM and the state-of-the-art methods, we adopt \textit{Frames Per Second} (FPS) as the evaluation metric. We further separately record the FPS change on our core components: CASC and ILMS modules. All experiments are conducted on an Nvidia 2080 GPU with an input size of 448$\times$448. As reported in Table \ref{time}, the inference speed of our method is lower than the baseline ResNet101 due to additional convolution operations involved in the ILMS module. Therefore, introducing ILMS or CASC module decreases FPS from 68.2 to 54.3 and 63.5, respectively. FPS further decreases to 52.4 when both modules are adopted. However, despite the decrease in speed, our method still achieves real-time performance. In addition, sacrificing only 15\% inference time improves mAP by 8\%. Our method also shows superior speed compared to SSGRL and comparable performance to ADD-GCD. This highlights the significant advantage of our SST in terms of the trade-off between both speed and accuracy.

\section{Conclusion}
In this paper, we propose a novel semantic-aware graph matching framework for multi-label image recognition, which reformulates the MLIR problem into a graph matching structure. By incorporating instance spatial graph and label semantic graph, and establishing instance-label assignment, the proposed ML-SGM utilizes graph network block to form structured representations for each instance and label by convolving its neighborhoods, which can effectively contribute the label dependencies and semantic-aware features to the learning model. We apply the proposed framework to MLIR, MLIR-PL and ML-FSL tasks, and extensive experimental results on various datasets significantly demonstrate its superiority over existing state-of-the-art methods on all three tasks. In summary, our proposed method is a multi-instance multi-label learning framework, and the selection of instance regions is crucial to the performance of multi-label image classification.
In future work, we will explore a more effective instance region generation module, which can well satisfy the high diversity, small number of instance regions, and high computational efficiency.

\section*{Acknowledgements} 
This work was supported by the Fundamental Research Funds for the Central Universities (No. 2022JBZY019).


\begin{IEEEbiography}[{\includegraphics[width=1in,height=1.25in,clip,keepaspectratio]{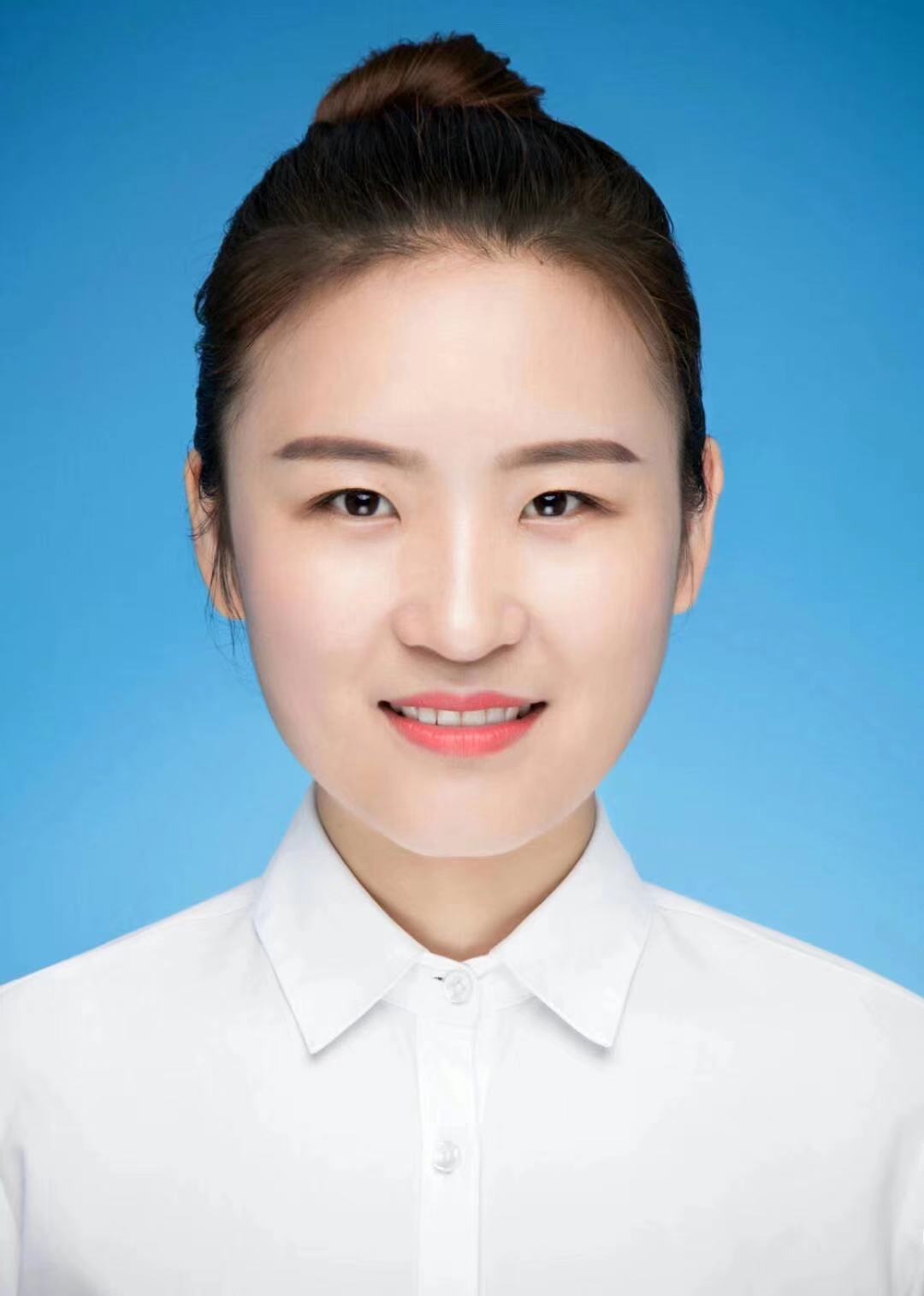}}]{Yanan Wu} received the B.Sc. degree from Henan University of Chinese Medicine, china, in 2016, and M.Sc. degree from the Department of Computer Science and Technology, Harbin University of Science and Technology, China, in 2019. she is currently pursuing the Ph.D. degree with the School of Computer and Information Technology, Beijing Jiaotong University. Her main research interests include computer vision and machine learning, especially in recognizing from multi-label image data.
\end{IEEEbiography}
 
\vspace{11pt}

\begin{IEEEbiography}[{\includegraphics[width=1in,height=1.25in,clip,keepaspectratio]{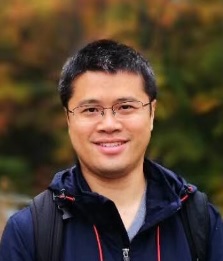}}]{Songhe Feng} received the BS degree and PhD degree from the School of Computer and Information Technology, Beijing Jiaotong University, Beijing, P.R. China, in 2003 and 2009, respectively. He is currently a full professor with the School of Computer and Information Technology, Beijing Jiaotong University. His research interests include machine learning and computer vision, especially in multi-view learning and weakly-supervised multi-label learning. He has published more than 60 peer-reviewed papers, including those in highly regarded journals and conferences such as the IEEE Trans. on Knowledge and Data Engineering, IEEE Trans. on Image Processing, IEEE Trans. on Cybernetics, IEEE Trans. on Multimedia, ACM Trans. on Knowledge Discovery from Data, AAAI, IJCAI, ACM SIGKDD, ECCV, ECML-PKDD, etc. He has been a visiting scholar with Michigan State University, USA and Dresden University of Technology, Germany, in 2014 and 2017, respectively.
\end{IEEEbiography}

\vspace{11pt}

\begin{IEEEbiography}[{\includegraphics[width=1in,height=1.25in,clip,keepaspectratio]{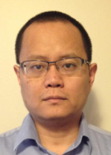}}]{Yang Wang} received the B.Sc. degree from the Harbin Institute of Technology, Harbin, China, the M.Sc. degree from the University of Alberta, Edmonton, AB, Canada, and the Ph.D. degree from Simon Fraser University, Burnaby, BC, Canada, all in computer science. He was previously a NSERC Postdoc Fellow with the University of Illinois at Urbana-Champaign, Champaign, IL, USA. He is currently an Associate Professor of computer science with the University of Manitoba, Winnipeg, MB, Canada. His research interests include computer vision and machine learning.
\end{IEEEbiography}


\end{document}